\documentclass[10pt,twocolumn,letterpaper]{article}

\usepackage{wacv}                      
\usepackage{CJKutf8}
\usepackage{amsmath,amssymb}
\usepackage{pifont}          
\usepackage[table]{xcolor}
\usepackage{float}
\usepackage{multirow}
\usepackage[normalem]{ulem}

\newcommand{\ourmethod}{SRVQ}

\definecolor{ours}{HTML}{68c4af}

\usepackage{amssymb}
\newcommand{\checkm}{{\color{ForestGreen}\ding{51}}}
\newcommand{\crossm}{{\color{red}\ding{55}}}

\definecolor{wacvblue}{rgb}{0.21,0.49,0.74}
\definecolor{hk}{rgb}{0.88,0.68,0.05}
\usepackage[breaklinks,colorlinks,allcolors=wacvblue]{hyperref}
\usepackage{booktabs}
\usepackage{multirow}
\usepackage{subcaption}
\usepackage{tikz}
\usetikzlibrary{positioning, arrows.meta, backgrounds, fit, patterns, calc}

\newcommand{\qualsamplethree}[3][0.99]{%
\begin{minipage}[t]{0.31\textwidth}
  \centering
  \parbox[c][2.5em][c]{0.98\linewidth}{\centering\small #3}
  \vspace{-1pt}

  \begin{minipage}{#1\linewidth}
    \makebox[0.18\linewidth][c]{\hspace{8pt}\footnotesize\bfseries GT}%
    \makebox[0.29\linewidth][c]{\hspace{14pt}\footnotesize\bfseries \shortstack{SRVQ\\sem.-only}}%
    \makebox[0.30\linewidth][c]{\footnotesize\bfseries \shortstack{SRVQ\\sem.+kin.}}%
    \makebox[0.22\linewidth][c]{\footnotesize\bfseries SOKE~\cite{soke}}%
  \end{minipage}%

  \vspace{1pt}
  \includegraphics[height=0.63\linewidth]{#2}
\end{minipage}%
}
\def\wacvPaperID{96}
\def\confName{WACV}
\def\confYear{2027}

\title{Semantic Residual Vector Quantization
       for Sign Language Generation}

\author{%
\setlength{\tabcolsep}{0pt}%
\begin{tabular}{c}
  \begin{tabular}{c}
    Chang Liu\textsuperscript{1}\\[1mm]
    chang.liu-2@unitn.it
  \end{tabular}
  \hspace{3mm}
  \begin{tabular}{c}
    Ke Han\textsuperscript{1,*}\\[1mm]
    ke.han.aca@gmail.com
  \end{tabular}
  \hspace{3mm}
  \begin{tabular}{c}
    Niccolò Biondi\textsuperscript{1}\\[1mm]
    niccolo.biondi@unitn.it
  \end{tabular}
  \hspace{3mm}
  \begin{tabular}{c}
    Gloria Isotton\textsuperscript{1}\\[1mm]
    gloria.isotton@unitn.it
  \end{tabular}
  \\[4mm]
  \begin{tabular}{c}
    Elisa Ricci\textsuperscript{1,2}\\[1mm]
    e.ricci@unitn.it
  \end{tabular}
  \hspace{5mm}
  \begin{tabular}{c}
    Davide Talon\textsuperscript{2}\\[1mm]
    dtalon@fbk.eu
  \end{tabular}
  \hspace{5mm}
  \begin{tabular}{c}
    Nicu Sebe\textsuperscript{1}\\[1mm]
    niculae.sebe@unitn.it
  \end{tabular}
  \\[2mm]
  \textsuperscript{1}University of Trento
  \quad
  \textsuperscript{2}Fondazione Bruno Kessler
\end{tabular}%
}

\begin{document}
\maketitle

\begingroup
\renewcommand{\thefootnote}{*}
\footnotetext{Corresponding author.}
\endgroup

\begin{abstract}
Sign language generation (SLG) maps written sentences into signing motion while preserving both lexical content and fine-grained articulation.
This requires a model to decide what signs should be expressed and to realize them through precise handshape, trajectory, facial expression and body coordination.
Existing pose-based systems and reconstruction-only motion tokenizers often entangle these two decisions, leaving the generator without an explicit text-aligned intermediate target for what to sign before refining how to sign it.
To address this issue, we introduce Semantic Residual Vector Quantization (SRVQ), a framework for SLG that frames generation as a two-level process moving from semantics to kinematics. The newly introduced SRV Quantizer learns a hierarchy in which the first residual level encodes sign content and subsequent levels refine pose, handshape, trajectory, and body coordination, all within a single quantizer and without separate semantic representations or external planning stages.
Through word-aware conditioning for hierarchical code generation, a masked semantic generator predicts semantic codes and a kinematic generator completes the residual codes for full SMPL-X motion synthesis.
We evaluate using retrieval and FID in a shared text-motion embedding space, alongside standard geometric metrics.
On CSL-Daily and Phoenix-2014T, SRVQ achieves R@1 of 82.3 and 70.3, outperforming the previous SOTA by 30 and 46 points, respectively.
\end{abstract}

\section{Introduction}
\label{sec:intro}

\begin{figure}[t]
  \centering
  \includegraphics[width=\columnwidth]{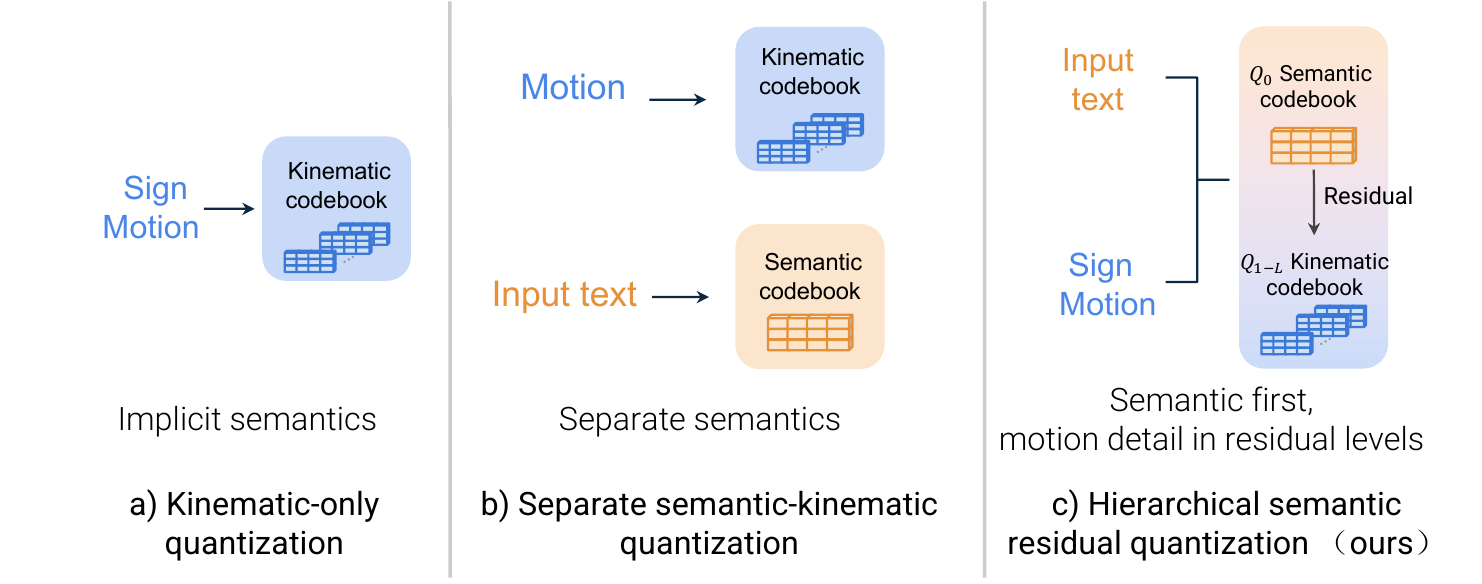}
\vspace{-6mm}
\caption{
Comparison of quantization paradigms for SLG.
(a) Kinematic-only quantization learns motion codes from motion alone, leaving semantics implicit.
(b) Separate semantic-kinematic quantization models semantics and motion separately using additional semantic representations.
(c) Our hierarchical semantic residual quantization embeds semantics and kinematics within a single residual quantizer, where the first residual level captures semantics and the remaining levels capture kinematic details.
}
\vspace{-3mm}
  \label{fig:teaser}
\end{figure}

Sign language is the primary language of many deaf and hard-of-hearing people, yet today’s digital society remains overwhelmingly built around spoken and written language~\cite{mack2020social,glasser2017deaf}. As a result, deaf users are often pushed toward text-based interaction rather than being supported in their native sign language. Sign language generation (SLG) aims to bridge this access gap by mapping written-language sentences into continuous signing motion, enabling more inclusive and accessible interaction.

SLG poses two coupled challenges: generating a signing sentence requires determining the linguistic content to express, \ie, which signs should be produced and in what order, while realizing each sign through fine-grained articulatory patterns, including handshapes, movement, facial expression, and body coordination. This makes SLG a two-level generation problem, entangling semantic sign selection with precise motion realization.
Yet, current models do not explicitly account for this two-level structure. They either generate signing motion end-to-end, mapping text directly to continuous poses~\cite{prog_transformer, adv_multichannel_slp, momp_slp}, leverage discrete motion tokens optimized for reconstruction-only~\cite{t2sgpt, m3t, soke}, or assume fixed intermediate sign representations that constrain the generation~\cite{spoken2sign}.
As a result, semantic sign selection and motion realization are modeled within a shared representation, making it difficult to learn representations tailored to their distinct roles.
Latent image generation~\cite{vqvae,maskgit} and text-to-motion models~\cite{t2mgpt,motiongpt,momask} suggest a broader principle: complex generation benefits from intermediate token spaces, where high-level content can be organized before low-level details are synthesized.
Recent models in these domains introduce semantic structure by learning reasoning tokens on top of pretrained motion tokenizers~\cite{think}, separating semantic and pixel branches in unified image tokenizers~\cite{qu2025tokenflow}, or first learning a semantic codebook and then adding low-level refinement codebooks through staged training~\cite{semhitok}.
These designs show the value of separating high-level semantics from low-level realization, but they usually require separate semantic spaces, staged training, or additional codebooks that are challenging to coordinate with reconstruction.

To address these gaps, we propose \textit{Semantic Residual Vector Quantization (SRVQ)}, a framework that formulates SMPL-X-based SLG in a \textit{semantic-to-kinematic} fashion, explicitly reflecting the two-level structure of signing, as illustrated in \cref{fig:teaser}.
At the core of SRVQ, the \textit{SRV Quantizer}  employs a residual quantization hierarchy where the first residual codebook is semantically aligned through text and gloss supervision to capture what should be signed, while subsequent residual codebooks progressively encode the fine-grained kinematic details needed to realize it. All within the same quantizer hierarchy. Generation follows the same decomposition: a \textit{semantic generator} predicts what to sign using word-aware conditioning that captures both global sentence and token-level lexical information, while a \textit{kinematic generator} completes how the signs are articulated. We evaluate SRVQ on CSL-Daily~\citep{csldaily} and Phoenix-2014T~\citep{phoenix} using retrieval in a shared text-motion space, together with standard geometric distances. SRVQ achieves 82.3 R@1 on CSL-Daily and 70.3 on Phoenix-2014T, outperforming SOKE~\citep{soke} by 30 and 46 points. Our contributions are summarized as follows:
\begin{itemize}
\item We introduce SRVQ, a semantic-to-kinematic generation framework that decomposes SLG into a two-level process that first predicts semantic information and then complement it with residual kinematics.

\item We present the SRV Quantizer, which encodes motion through a single residual hierarchy learned with both reconstruction and text-alignment objectives. The first-level codebook captures high-level semantics, while later residual levels encode fine-grained kinematic details.

\item We introduce word-aware conditioning, in which both generators cross-attend to word-token embeddings augmented with a global sentence token, giving motion tokens direct access to local lexical cues while retaining sentence-level context.

\item SRVQ achieves state-of-the-art performance in SLG as measured with both retrieval and geometric metrics. 

\end{itemize}

\begin{figure*}[t]
  \centering
  \includegraphics[width=1.0\textwidth]{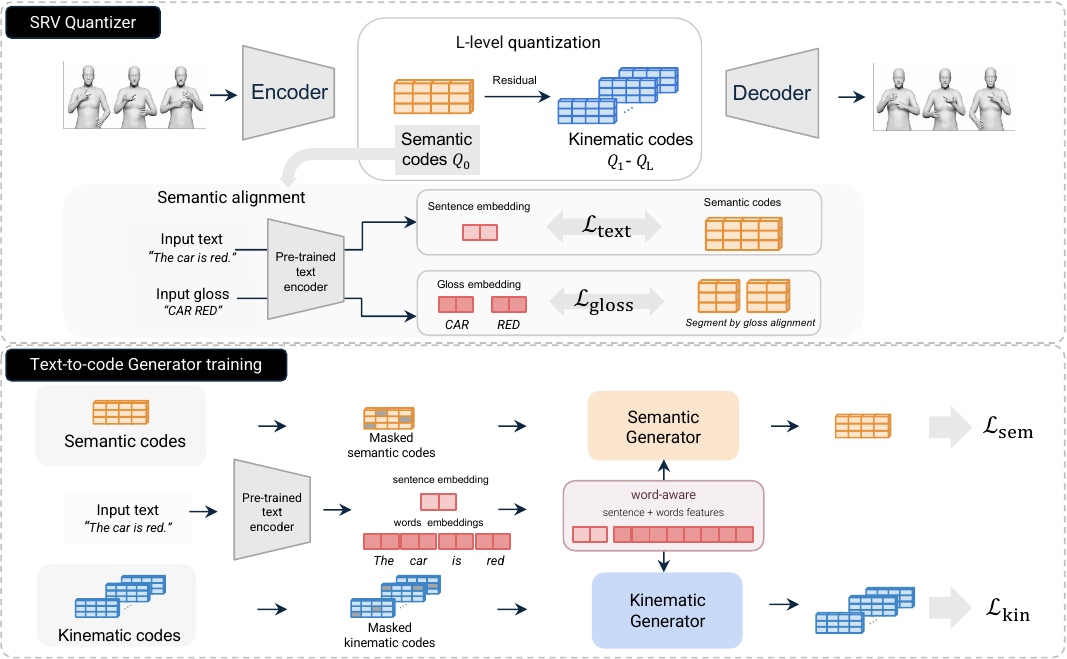}
\caption{\textbf{Overview of the SRVQ framework.}
\textit{Top:} 
The SRV Quantizer learns a hierarchical discrete representation of each SMPL-X motion.
The first residual level ($Q_0$) learns a text-aligned semantic codebook, while the remaining residual levels $Q_1$--$Q_L$ progressively capture finer kinematic details. \textit{Bottom:} During generation, word embeddings together with a global sentence token form a word-aware textual representation. A Semantic Generator predicts $Q_0$ from this representation, and a kinematic generator predicts $Q_1$-$Q_L$ before the part decoders reconstruct the continuous SMPL-X motion.}
\vspace{-2mm}
  \label{fig:pipeline}
\end{figure*}
\section{Related Work}
\label{sec:related}

\noindent\textbf{Sign Language Generation.}
SLG aims to synthesize continuous signing motion from text input, and prior work has explored several ways to bridge this linguistic-to-kinematic gap.
Progressive Transformers~\cite{prog_transformer} directly regress skeletal motion from text, while adversarial multi-channel training~\cite{adv_multichannel_slp} and motion-primitive modeling~\cite{momp_slp} improve continuous pose generation.
Other methods synthesize human appearance through 3D avatars or meshes~\cite{text2mesh,neural_sign_actors}, while related controllable image-generation methods introduce explicit pose or appearance conditions~\cite{imagpose,imagdressing}. Diffusion-based methods model signing motion directly~\cite{signdiff}.
Recent discrete approaches introduce a motion vocabulary before generation.
T2S-GPT~\cite{t2sgpt} uses dynamic vector quantization for autoregressive sign generation, SOKE~\cite{soke} combines discrete sign tokens with retrieval-enhanced conditioning, and M3T~\cite{m3t} introduces discrete multi-modal motion tokens. Several of these systems additionally use gloss annotations as intermediate supervision~\cite{prog_transformer, xie2024g2p}, where glosses provide word-like labels for individual sign units and serve as sign-level anchors between spoken-language text and signing motion.
These methods improve sequence modeling, but the tokenizer is usually trained mainly for motion reconstruction.
As a result, the discrete code space is not explicitly organized to separate sign content from kinematic realization.
This motivates us to look beyond SLG to tokenized text-to-image and motion generation, where discrete or latent representations bridge high-level content and low-level synthesis.

\noindent\textbf{Semantic Organization in Tokenized Generation.}
Vector quantization provides a discrete interface between continuous signals and sequence models~\cite{vqvae}.
In text-to-image generation, masked and autoregressive token models predict visual tokens before decoding them into pixels~\cite{maskgit,rqtransformer,ctf}, suggesting that an intermediate token space can separate high-level content prediction from low-level synthesis.
Text-to-motion generation adopts a similar tokenized formulation for temporally extended human motion.
T2M-GPT~\cite{t2mgpt} and MotionGPT~\cite{motiongpt} model motion as discrete token sequences, while later methods further explore semantic-to-kinematic separation through global-local attention~\cite{attt2m}, text-aligned motion representations~\cite{humantomato}, explicit planning and posture stages~\cite{promotion,text2bfm}, or reasoning and execution latents~\cite{think}.
These works indicate that generation benefits from exposing an intermediate semantic structure before producing detailed motion.
However, this semantic-reconstruction balance is often introduced through additional objectives, separate representations, or multi-stage generator designs~\cite{attt2m,humantomato,promotion,text2bfm,think}.
Residual quantization~\cite{rqtransformer,soundstream,encodec,momask} offers a natural hierarchy of codebooks, where later levels progressively reduce the residual error left by earlier ones. In existing image, audio, and text-to-motion tokenizers, this hierarchy is mainly used to improve reconstruction fidelity or rate-distortion trade-offs. SRVQ instead uses the hierarchy as a semantic-to-kinematic interface for SLG, assigning the first residual level to text-aligned sign content and the later levels to kinematic refinement.

\noindent\textbf{Text Conditioning for Sign Generation.}
Beyond the tokenizer, the text representation used by the generator also matters.
Many SLG systems inherit the text-to-motion practice of conditioning generation on a pooled sentence embedding or a small set of global text tokens~\cite{tevet2022human, liu2023spatio, t2sgpt,soke}.
This design gives the generator sentence-level context, but it can hide the local correspondences between words and sign segments.
The limitation is especially important for SLG because individual lexical units often determine handshape, spatial location, handedness, and local motion timing.
Our generator instead predicts semantic motion codes by cross-attending to word-token embeddings augmented with a global sentence token.
This exposes local lexical cues to every generated motion token while retaining sentence-level context for the whole sequence.


\section{Method}
\label{sec:method}
We introduce \ourmethod{}, a semantic-to-kinematic framework for SLG, as shown in \cref{fig:pipeline}. It first learns a \textit{SRV Quantizer} (\cref{fig:pipeline}-top, \cref{sec:tokenizer}) that decomposes signing motion into a text-aligned semantic level and residual kinematic levels. Later, in the \textit{Word-Aware Conditioning for Hierarchical Code Generation} (\cref{fig:pipeline}-bottom, \cref{sec:generation}), we train a semantic generator to predict semantic codes ($Q_0$) from text, while a kinematic generator completes the residual codes ($Q_1$--$Q_L$) required for full motion synthesis.

\noindent\textbf{Problem Formulation.} Given an input sentence $\mathbf{c}=(c_1,\ldots,c_M)$, where $c_i$ denotes the $i$-th word, SLG aims to generate a continuous SMPL-X signing motion sequence $\mathbf{S}\in\mathbb{R}^{T\times133}$ that preserves the sentence meaning.
Following~\cite{soke}, we represent $\mathbf{S}$ using a 133-dimensional SMPL-X parameterization and decompose it into body (including facial features), left-hand, and right-hand streams, denoted by $\mathbf{S}^{B}$, $\mathbf{S}^{L}$, and $\mathbf{S}^{R}$.
We first describe how the SRV Quantizer learns a motion vocabulary whose first code level is aligned with sign content, and then introduce the hierarchical generators that predict semantic and residual codes from text.

\subsection{SRV Quantizer}
\label{sec:tokenizer}
\begin{figure}[t]
  \centering
  \includegraphics[width=\linewidth]{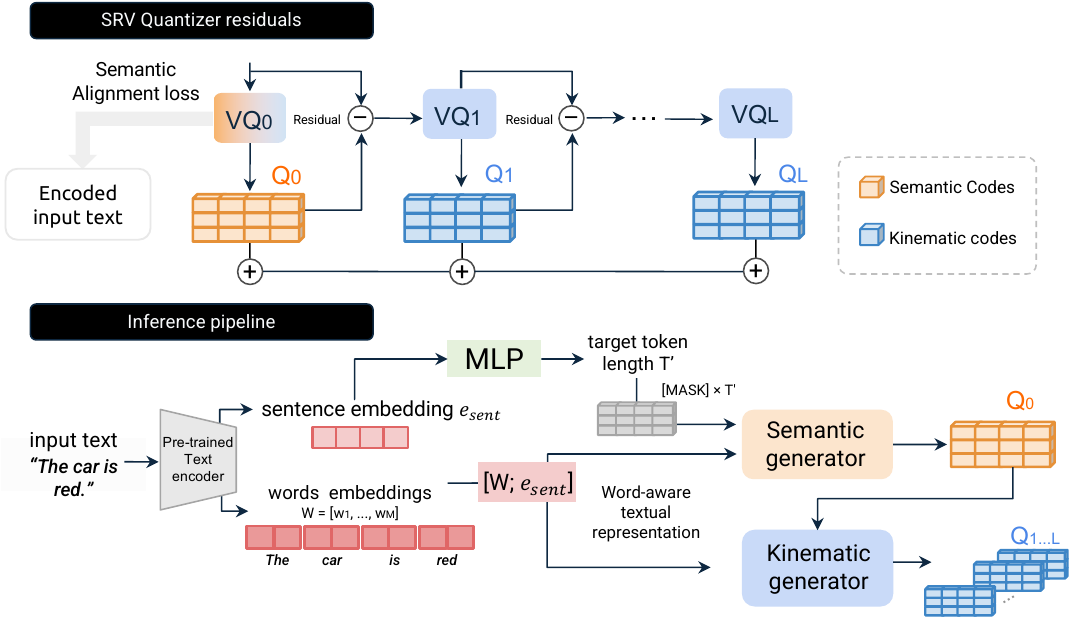}
\caption{
\textit{Top:} Each quantization level encodes the residual from the previous level, with semantic supervision applied only to $Q_0$. The final representation consists of the semantic code and residual kinematic codes.
\textit{Bottom:} During inference, the Semantic Generator predicts $Q_0$ from the word-aware textual representation, followed by the Kinematic Generator predicting $Q_1,\ldots,Q_L$.
}
  \label{fig:rvq}
\end{figure}

Residual Vector Quantization (RVQ)~\citep{soundstream,encodec} provides a natural mechanism for learning hierarchical discrete representations. We exploit this structure to separate semantics, \ie, what to sign, from kinematic refinement, \ie, how to sign it, within a single end-to-end quantizer, without imposing explicit constraints on how modeling capacity is allocated across levels. Specifically, the first residual level is trained as a text-aligned semantic codebook, capturing the meaning of the sign content to be produced, while the subsequent levels encode residual motion details that specify how the sign is articulated. 
This design keeps semantic code prediction and motion reconstruction within the same hierarchy, providing the generator with an explicit semantic target while subsequent levels capture fine-grained motion details.

The SRV Quantizer processes $\mathbf{S}^{B}$, $\mathbf{S}^{L}$, and $\mathbf{S}^{R}$ independently, with each body-part using its own encoder/decoder and codebooks. For clarity, we describe the quantizer for a generic stream $\mathbf{X}=\mathbf{S}^p, p\in\{B, L, R\}$ and omit the part superscript.
For each body part, a 1D-CNN encoder maps the motion $\mathbf{X}$ to a latent sequence $\mathbf{F}\in\mathbb{R}^{T'\times d}$, with temporal downsampling factor $u=4$. We apply an $L+1$-level residual quantizer to $\mathbf{F}$. Starting from the initial residual sequence $\mathbf{r}^{0}=\mathbf{F}$, the quantizer recursively approximates the residual at each level and passes the remaining residual to the next level. At level $\ell$, each residual token is assigned to its nearest entry in a learned codebook $\mathbf{Z}^{\ell}=\{\mathbf{z}_{n}^{\ell}\}_{n=1}^{N^\ell}$:
\begin{equation}
\begin{aligned}
i_t^\ell &= \arg\min_n \left\| \mathbf{r}_t^\ell-\mathbf{z}_n^\ell \right\|_2^2, 
\qquad 
\hat{\mathbf{z}}_t^\ell = \mathbf{z}_{i_t^\ell}^{\ell},\\
\hat{\mathbf{Z}}^\ell &= 
\left(\hat{\mathbf{z}}_1^\ell,\ldots,\hat{\mathbf{z}}_{T'}^\ell\right),
\qquad
\mathbf{r}^{\ell+1} = \mathbf{r}^{\ell} - \hat{\mathbf{Z}}^\ell .
\end{aligned}
\label{eq:rvq}
\end{equation}
Here, $\mathbf{r}^{\ell}\in\mathbb{R}^{T'\times d}$ denotes the residual sequence at level $\ell$, $\mathbf{r}^{\ell}_{t}\in\mathbb{R}^{d}$ denotes its token at temporal position $t$, and $\hat{\mathbf{Z}}^\ell\in\mathbb{R}^{T'\times d}$ is the quantized sequence selected at level $\ell$.
We denote the level-$\ell$ discrete index sequence by
$Q_\ell=(i_1^\ell,\ldots,i_{T'}^\ell)$; looking up these indices in the
codebook $\mathbf{Z}^{\ell}$ yields the quantized embeddings
$\hat{\mathbf{Z}}^\ell$.

After the final residual level, the decoder $\mathcal{D}$  reconstructs the motion stream from the sum of the selected code embeddings: $\hat{\mathbf{X}}=
\mathcal{D}
\left(
\sum_{\ell=0}^{L}
\hat{\mathbf{Z}}^\ell
\right)$.

We encourage level-0 codes to model semantic information, while leaving the remaining residual levels free to capture fine-grained kinematic details (\cref{fig:rvq}-top). To this end, we concatenate the embeddings from the three part branches $\hat{\mathbf{Z}}^{p,0}$ and project them into a text-motion $d$-dimensional contrastive space:
\begin{equation}
\mathbf{H}^{0} =
[\hat{\mathbf{Z}}^{B,0};
 \hat{\mathbf{Z}}^{L,0};
 \hat{\mathbf{Z}}^{R,0}], \quad \mathbf{M} = \mathrm{norm}(\phi(\mathbf{H}^{0})) \in \mathbb{R}^{T'\times d},
\label{eq:semantic_projection}
\end{equation}
with $\phi$ denoting a two-layer MLP.
A pretrained text encoder~\cite{labse} produces word-token embeddings for the input sentence $\mathbf{c}$, which are projected to the same $d$-dimensional contrastive space as $\mathbf{M}$, yielding
$\mathbf{W}=(\mathbf{w}_1,\ldots,\mathbf{w}_M)$. Here, $\mathbf{M}$ is a continuous feature for semantic alignment and is
distinct from the discrete index sequence $Q_0$.
Let $\operatorname{sim}(\cdot,\cdot)$ denote cosine similarity. We define a bidirectional token-level matching score between motion features $\mathbf{M}$ and word tokens $\mathbf{W}$ as:
\begin{equation}
\begin{aligned}
\operatorname{match}(\mathbf{M}, \mathbf{W})
&=
\frac{1}{2}
\bigg(
\frac{1}{T'}\sum_{t=1}^{T'}\max_{1\le i\le M}\text{sim}(\mathbf{m}_t, \mathbf{w}_i)
\\
&\quad+
\frac{1}{M}\sum_{i=1}^{M}\max_{1\le t\le T'}\text{sim}(\mathbf{m}_t, \mathbf{w}_i)
\bigg),
\end{aligned}
\label{eq:token_sim}
\end{equation}
where $\mathbf{m}_t$ and $\mathbf{w}_i$ denote the $t$-th row of $\mathbf{M}$ and the $i$-th row of $\mathbf{W}$, respectively, and the first term measures how well each motion feature is explained by its most similar word token, while the second term measures how well each word token is covered by the motion sequence. We use this matching score in an InfoNCE loss~\citep{oord2018representation} over the samples in each mini-batch, encouraging the level-$0$ motion features to align with their paired text while contrasting against others:
\begin{equation}
\label{eq:text_infonce}
\mathcal{L}_{\mathrm{text}}
=
-\frac{1}{|\mathcal{B}|}
\sum_{j\in\mathcal{B}}
\log
\frac{
\exp(\operatorname{match}(\mathbf{M}_j,\mathbf{W}_j)/\tau)
}{
\sum_{k\in\mathcal{B}}
\exp(\operatorname{match}(\mathbf{M}_j,\mathbf{W}_k)/\tau)
},
\end{equation}
with $\mathcal{B}$ the mini-batch and $\tau$ the contrastive temperature.

Similarly, we use frame-level gloss annotations to provide local semantic supervision for the level-0 motion features $\mathbf{M}$.
Each gloss label is associated with a start and end frame in the motion sequence, which we use to crop the corresponding segment of $\mathbf{M}$.
Let $(\mathbf{M}_i, \mathbf{G}_i)$ denote the cropped motion-feature segment and its associated gloss text features in a mini-batch, where $\mathbf{G}_i$ is obtained by encoding the gloss label with the same text encoder~\cite{labse}.
We define $\mathcal{L}_{\mathrm{gloss}}$ as the same InfoNCE objective as Eq.~\ref{eq:text_infonce}, replacing full motion-text pairs $(\mathbf{M}_j,\mathbf{W}_j)$ with gloss cropped-motion ones.

For motion reconstruction, we use a Smooth-L1 loss $\mathcal{L}_{\mathrm{rec}}$ between reconstructed and ground-truth motion streams.
Following prior motion tokenization practice~\cite{momask}, we add the first-order velocity loss $\mathcal{L}_{\mathrm{vel}}$ on $\Delta\mathbf{X}_t=\mathbf{X}_{t}-\mathbf{X}_{t-1}$ to encourage temporal smoothness. We also use the standard VQ commitment loss $\mathcal{L}_{\mathrm{commit}}$~\cite{vqvae,soundstream,encodec}, which keeps encoder residuals close to their assigned code embeddings.
The final objective for the SRV Quantizer is:
\begin{align} 
\label{eq:tokenizer_objective}
\mathcal{L}
={}&
\mathcal{L}_{\mathrm{rec}}
+\lambda_{\mathrm{text}}\mathcal{L}_{\mathrm{text}}
+\lambda_{\mathrm{gloss}}\mathcal{L}_{\mathrm{gloss}}\\
&+\lambda_{\mathrm{vel}}\mathcal{L}_{\mathrm{vel}}
+\lambda_{\mathrm{commit}}\mathcal{L}_{\mathrm{commit}} \nonumber,
\end{align}
with $\lambda_\text{text}, \lambda_\text{gloss}, \lambda_\text{vel}$ and $\lambda_\text{commit}$ hyperparameters regulating the strengths of the respective losses.

\subsection{Word-Aware Conditioning for Hierarchical Motion Generation}
\label{sec:generation}
In the second stage, we freeze the SRV Quantizer and train generators to predict its discrete code hierarchy from text.
The quantizer converts each motion sequence into part-level code streams: $Q_0$ codes represent text-aligned sign content, while higher residual levels represent kinematic detail.
Generation follows the same hierarchy. A semantic generator first predicts the $Q_0$ codes, and a kinematic generator then completes the residual levels for reconstruction.

\noindent\textbf{Word-Aware Conditioning.}
To condition generation on both local lexical information and sentence-level context, we encode the input sentence $\mathbf{c}$ with a pretrained text encoder~\cite{labse}.
This produces word-token embeddings $\mathbf{W}$ and a global sentence embedding $\mathbf{e}_{\mathrm{sent}}$.
We append the sentence embedding to the word-token sequence and use the resulting word-aware textual representation for conditioning generation via cross-attention.
This lets each motion token attend to local lexical cues while retaining the sentence-level context of the full sentence.

\noindent\textbf{Semantic Generator.}
The semantic generator predicts $Q_0$ with a bidirectional masked transformer~\cite{maskgit,momask}.
During training, we randomly mask a subset of $Q_0$ tokens with a learnable \texttt{[MASK]} token and train the model to recover the masked code indices by cross-attending to the word-aware textual representation. We rely on a cross-entropy loss over masked positions:
\begin{equation}
\mathcal{L}_{\mathrm{sem}}=
-\frac{1}{|\mathcal{I}|}\sum_{j\in\mathcal{I}}
\log p\left(i^{0}_{j}\mid Q_{0,\setminus \mathcal{I}}, \mathbf{W}, \mathbf{e}_{\mathrm{sent}}\right),
\end{equation}
where $i_j^0$ is the masked level-0 code index at position $j$ and $Q_{0,\setminus \mathcal{I}}$ denotes the visible level-0 indices. Since each body part is quantized with a separate codebook, we predict all body-part codes jointly using separate part heads.

\noindent\textbf{Kinematic Generator.}
The Kinematic Generator follows the same masked prediction formulation, but operates on the residual levels rather than the $Q_0$ semantic codes. We denote the summed cross-entropy loss over residual levels $1$ to $L$ as $\mathcal{L}_{\mathrm{kin}}$.
Training leverages a teacher-forcing strategy where lower-level codes are obtained by quantization of the ground-truth motion.

\noindent\textbf{Inference.}
As shown in Fig.~\ref{fig:rvq}-bottom, at inference time, a lightweight MLP first predicts the target token length $T'$ from the global sentence embedding $\mathbf{e}_{\mathrm{sent}}$. The length predictor is trained with cross-entropy over discretized length bins, using the 
length of the ground-truth motion as supervision. Given the predicted length, the semantic generator starts from a fully masked sequence of length $T'$, to progressively decode semantic codes using confidence-based unmasking~\cite{maskgit}. To improve text adherence, we apply classifier-free guidance (CFG)~\cite{cfg} during decoding. The resulting semantic codes are then passed to the kinematic generator, which predicts the residual codes needed to recover fine-grained articulatory details.

\section{Experiments}
\label{sec:experiments}
We evaluate \ourmethod{} on standard sign language generation benchmarks. We first describe the evaluation protocol, including implementation details, datasets, and metrics. We then compare against prior work through both quantitative and qualitative evaluations, followed by ablation studies that isolate the contribution of the key design choices.

\begin{table*}[t]
\centering
\caption{\textbf{Comparison with state-of-the-art methods on CSL-Daily and Phoenix-2014T.} The best and second-best results are highlighted in \textbf{bold} and \underline{underline}, respectively.}

\label{tab:main}
\setlength{\tabcolsep}{3.4pt}
\resizebox{\textwidth}{!}{%
\begin{tabular}{l cccccccc cccccccc}
\toprule
 & \multicolumn{8}{c}{\textbf{CSL-Daily}} & \multicolumn{8}{c}{\textbf{Phoenix-2014T}} \\
\cmidrule(lr){2-9}\cmidrule(lr){10-17}
\textbf{Method}
 & \multicolumn{4}{c}{Retrieval / FID}
 & \multicolumn{2}{c}{DTW-JPE}
 & \multicolumn{2}{c}{DTW-PA}
 & \multicolumn{4}{c}{Retrieval / FID}
 & \multicolumn{2}{c}{DTW-JPE}
 & \multicolumn{2}{c}{DTW-PA} \\
 \cmidrule(lr){2-5}
 \cmidrule(lr){6-7}
 \cmidrule(lr){8-9}
 \cmidrule(lr){10-13}
 \cmidrule(lr){14-15}\cmidrule(lr){16-17}
 & R@1$\uparrow$ & R@5$\uparrow$ & R@10$\uparrow$ & FID$\downarrow$
 & Body$\downarrow$ & Hands$\downarrow$ & Body$\downarrow$ & Hands$\downarrow$
 & R@1$\uparrow$ & R@5$\uparrow$ & R@10$\uparrow$ & FID$\downarrow$
 & Body$\downarrow$ & Hands$\downarrow$ & Body$\downarrow$ & Hands$\downarrow$ \\
\midrule
Prog. Trans.~\cite{prog_transformer}
 & -- & -- & -- & -- & 16.30 & 32.63 & 15.98 & 12.91
 & -- & -- & -- & -- & 15.01 & 31.77 & 13.67 & 11.95 \\
Text2Mesh~\cite{text2mesh}
 & -- & -- & -- & -- & 13.76 & 30.37 & 13.47 & 12.10
 & -- & -- & -- & -- & 14.04 & 31.64 & 13.48 & 12.06 \\
T2S-GPT~\cite{t2sgpt}
 & -- & -- & -- & -- & 12.32 & 15.43 & 11.94 & 5.93
 & -- & -- & -- & -- & 11.65 & 19.09 & 10.38 & 6.47 \\
S-MotionGPT~\cite{motiongpt}
 & -- & -- & -- & -- & 11.58 & 11.31 & 10.81 & 3.78
 & -- & -- & -- & -- & 10.42 & 9.08 & 9.45 & 3.41 \\
SOKE~\cite{soke}
 & 51.96 & 69.30 & 75.77 & 0.062 & \underline{6.541} & \underline{8.524} & \underline{7.407} & \textbf{1.503}
 & 24.14 & 45.64 & 54.67 & 0.151 & 5.914 & 7.316 & 6.464 & 1.368 \\
\rowcolor{ours!20}SRVQ (semantic-only)
 & \underline{77.81} & \underline{89.97} & \underline{93.28} & \textbf{0.057}
 & \textbf{6.296} & \textbf{8.376} & \textbf{7.113} & \underline{1.508}
 & \underline{57.48} & \underline{78.97} & \underline{85.83} & \underline{0.083}
 & \textbf{4.700} & \textbf{6.410} & \textbf{5.315} & \textbf{1.262} \\
\rowcolor{ours!20}SRVQ (sem+kinematic)
 & \textbf{82.31} & \textbf{94.13} & \textbf{96.34} & \underline{0.058}
 & 6.787 & 8.535 & 7.614 & 1.544
 & \textbf{70.25} & \textbf{87.54} & \textbf{92.52} & \textbf{0.059}
 & \underline{5.330} & \underline{6.920} & \underline{6.003} & \underline{1.354} \\
\bottomrule
\end{tabular}
\vspace{-3mm}
}
\end{table*}

\subsection{Experimental Setup}
\label{sec:impl}

\noindent\textbf{Implementation Details.} SRVQ is trained jointly on CSL-Daily and Phoenix-2014T.
It uses 6 quantization levels ($L=5$) with code dimension 512, body codebook size $N^{\ell}_B = 96$, and hand codebook sizes $N^{\ell}_{L} = N^{\ell}_{R} = 256$ at each level.
The quantizer loss of Eq.~\ref{eq:tokenizer_objective} uses $\lambda_{\mathrm{text}} = 0.1$, $\lambda_{\mathrm{gloss}} = 0.1$, $\lambda_{\mathrm{vel}}=0.5$ and $\lambda_{\mathrm{commit}}=0.02$. 
The contrastive temperature of Eq. \ref{eq:text_infonce} is $\tau = 0.07$, $d=512$. 
Both the semantic and kinematic generators are 8-layer masked Transformers.
We train all modules with AdamW~\citep{loshchilovdecoupled} using learning rate $2\times10^{-4}$ and batch size 64 per GPU on 4 GPUs.
The quantizer is trained for 200 epochs, while the two generators are trained for 300.
Evaluation is conducted on CSL-Daily and Phoenix-2014T test sets.
At inference, we use CFG scale $\gamma = 2.5$, and $K = 10$ decoding steps. Unless otherwise stated, we use inference seed 3407, following the default configuration of MoMask~\citep{momask}.

\noindent\textbf{Datasets.} We evaluate on CSL-Daily~\cite{csldaily} (18K training / 1176 test sentences, Chinese sign language) and Phoenix-2014T~\cite{phoenix} ($\sim$7K training / 642 test sentences, German sign language). Following prior work~\cite{soke}, we use the released 133-dimensional SMPL-X pose annotations sampled at 25 fps. For gloss supervision, we rely on the frame-level annotations from Spoken2Sign~\cite{spoken2sign}, which provide gloss labels together with their start and end frames.

\noindent\textbf{Evaluation Metrics.} Following prior  works~\cite{prog_transformer,soke}, we report DTW-JPE and DTW-PA.
DTW first aligns generated and reference motions over time with dynamic time warping, which reduces the effect of length and speed mismatch.
DTW-JPE then computes the average joint position error over the aligned frame pairs.
DTW-PA additionally applies Procrustes alignment before computing the same joint error, so it discounts global translation, rotation, and scale differences.
Lower values indicate a smaller geometric distance to the reference motion and, therefore, better fidelity.
However, geometric metrics alone are insufficient for assessing semantic correctness. A sentence can be signed with valid variations in timing, coarticulation, and style, and imperfect spatiotemporal alignment further complicates reference-based evaluation. As a result, geometric distance to a single reference cannot fully capture whether the generated motion preserves the intended meaning. Recent SLG evaluation studies therefore advocate embedding-based semantic evaluation as a complement to pose-distance metrics~\cite{silverscore,meaningful_pose_eval}. Retrieval-based semantic matching is also widely used in text-to-motion generation~\cite{t2mgpt,tmr,momask,party} and has recently been adopted in sign-language generation~\cite{yu2024signavatars,bensabath2026handmdm}. We use retrieval metrics as our primary semantic indicator. To this end, a pretrained text-motion retrieval model~\cite{retrieval} embeds generated motions and text into a shared space.
For each benchmark, this evaluator is trained on real SMPL-X motion-text pairs from the corresponding training split and is used for all generated and real motions.
We report motion-to-text $R@1$, $R@5$, $R@10$, and FID. 
We refer to the Supplementary~\ref{supp:additional-eval} for additional validation using Meaningful Pose and back-translation evaluators adapted to SMPL-X motion.

\subsection{Main Results}
\label{sec:sota}

\noindent\textbf{Baselines.} We compare \ourmethod{} against representative sign language generation baselines reported by SOKE~\cite{soke}. These include Progressive Transformer (Prog.~Trans.)~\cite{prog_transformer}, which directly regresses continuous signing poses; Text2Mesh~\cite{text2mesh}, which leverages mesh-based generation; and discrete motion-token generation methods, namely T2S-GPT~\cite{t2sgpt} and S-MotionGPT~\cite{motiongpt} (MotionGPT~\cite{motiongpt} adapted for sign language as evaluated by SOKE). To ensure comparability, we report geometric metrics using the numbers provided by SOKE~\cite{soke}. Since prior work does not report retrieval-based metrics, we additionally evaluate SOKE under the same retrieval protocol used for \ourmethod{}.

\noindent\textbf{Quantitative Results.} \cref{tab:main} compares \ourmethod{} with state-of-the-art SLG models. We report two variants: \emph{Semantic-only}, which decodes motion after predicting only the level-0 semantic codes $Q_0$, and \emph{sem+kinematic}, which additionally predicts the residual kinematic codes before decoding. 

\ourmethod{} outperforms prior methods across nearly all considered metrics. Sem+kinematic achieves the strongest semantic retrieval performance on both datasets, while semantic-only performs best on most geometric metrics. 

Intuitively, semantic-only produces a coarse, temporally smooth, low-frequency representation that is close to the average of many valid articulations of the same sign, which is well aligned with what DTW-based metrics reward. We revisit this observation in Sec.~\ref{sec:length}. In contrast, sem+kinematic adds articulation details, improving the model's ability to capture fine-grained distinctions between samples. On CSL-Daily, sem+kinematic improves R@1 from 51.96 to 82.31, yielding a gain of 30.35 percentage points over the strongest prior baseline. On Phoenix-2014T, it improves R@1 from 24.14 to 70.25, a gain of 46.11 points, and also obtains the best FID. 
Notably, most of the semantic retrieval gain already comes from explicit semantic modeling, supporting the proposed hierarchical semantic modelling. 
Retrieval improvements are consistent across R@5 and R@10, indicating that the generated motions are not only ranked closer to the correct sentence at top-1, but also occupy a better neighborhood in the embedding space. 
For reference, the retrieval R@1 of real test motions under the same protocol is 87.0 on CSL-Daily and 71.0 on Phoenix-2014T, showing that Sem+kinematic substantially closes the gap to ground-truth retrievability.



\noindent\textbf{Qualitative Results.}
\begin{figure*}[t]
  \centering
  \setlength{\tabcolsep}{2pt}
  \renewcommand{\arraystretch}{0.85}
  \resizebox{\textwidth}{!}{%

  \begin{tabular}{ccc}
    \qualsamplethree[0.99]{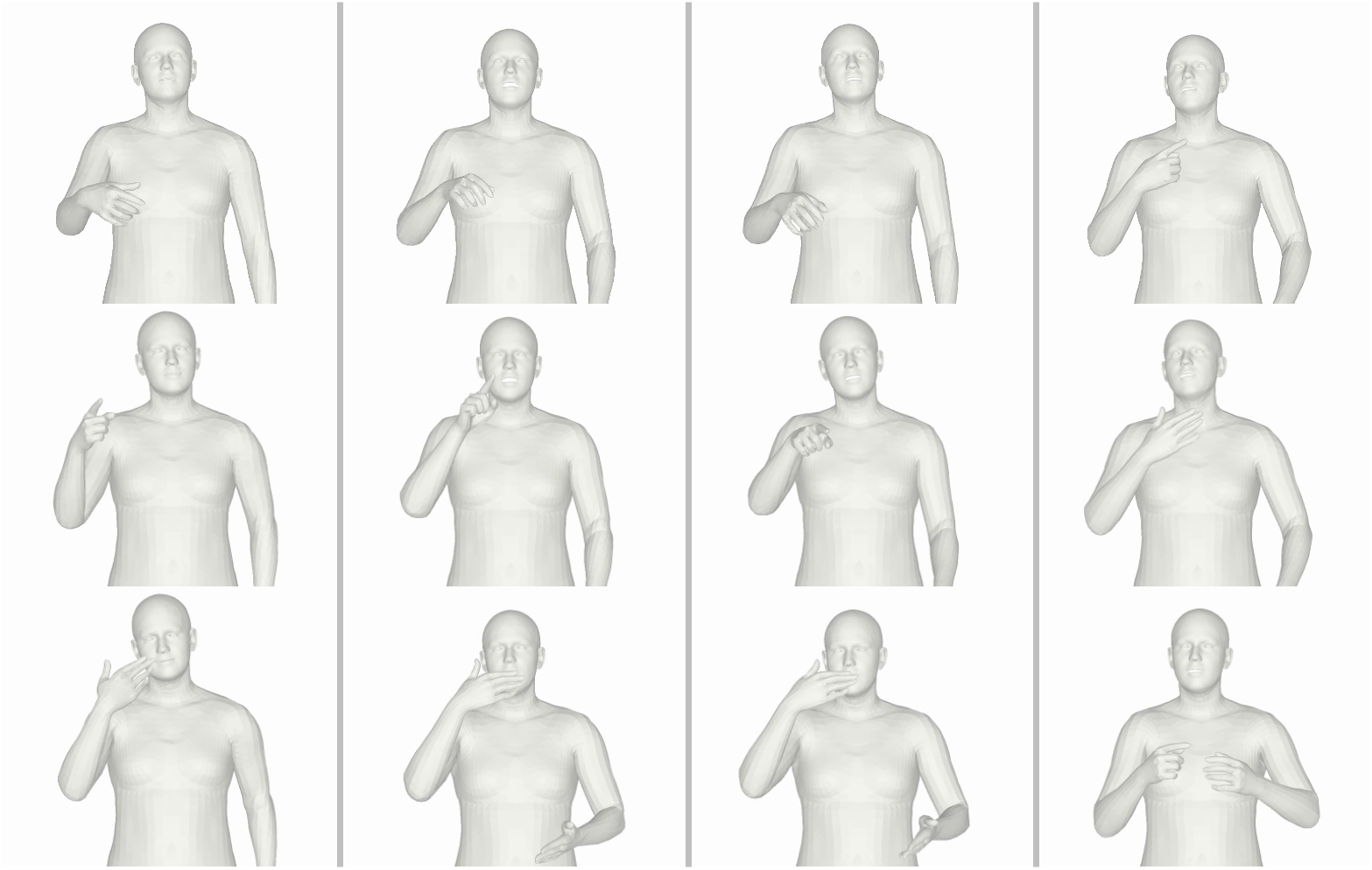}
      {You’re my client, so dinner’s on me.}
    &
    \qualsamplethree[0.99]{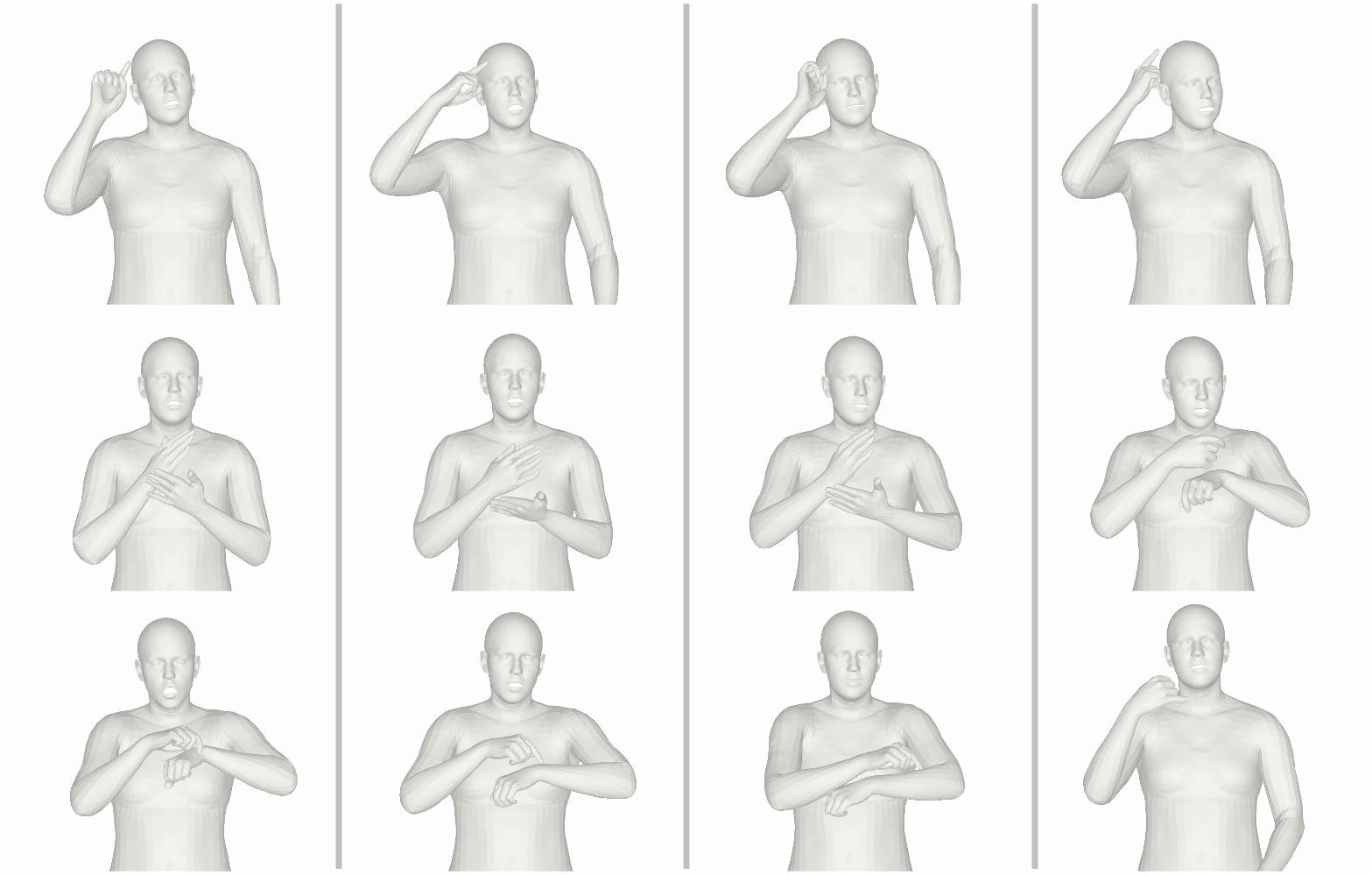}
      {Exam tomorrow. Bring a pen, no phone.}
    &
    \qualsamplethree[0.92]{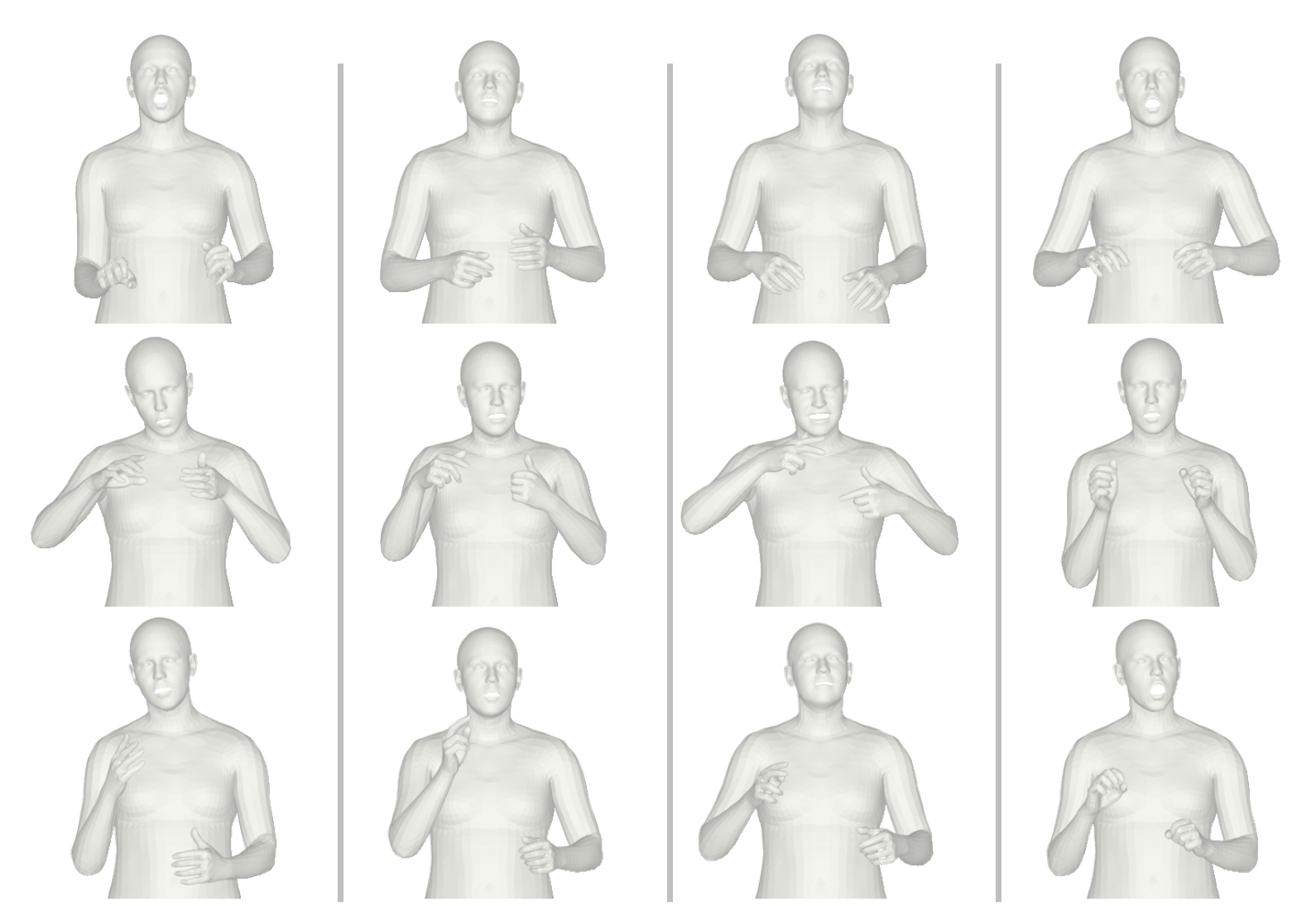}
      {Today's weather will definitely remain wintry.}
  \end{tabular}
  }
  \caption{\textbf{Qualitative comparison of generated sign sequences.}
  We show representative frames from ground truth (GT), SRVQ semantic-only generation,
  SRVQ sem+kinematic generation, and SOKE. The text above each example is the
  corresponding input sentence. The first two examples are from CSL-Daily, and
  the last example is from Phoenix-2014T.}
  \label{fig:qual}
\vspace{-3mm}
\end{figure*}
Fig.~\ref{fig:qual} visualizes representative generated sequences alongside the ground-truth signing motion.
Semantic-only generation already recovers sentence-dependent signing patterns, showing that the semantic codes provide a meaningful text-aligned target for generation.
Sem+kinematic preserves this semantic structure while adding residual hand and body details, producing motions that are locally coherent over the sampled sequence.
Compared with SOKE, SRVQ yields sign sequences that better follow the intended lexical content, which is consistent with the retrieval gains in \cref{tab:main}.

\subsection{Ablation Studies}
\label{sec:ablation}

    \noindent\textbf{Semantic Alignment.} \cref{tab:lambda} examines the effect of semantic supervision on the quantizer. Results show that removing both text and gloss alignment ($\lambda_{\text{text}}=0,\lambda_{\text{gloss}}=0$) yields the worst performance across most metrics, highlighting the importance of injecting semantic information into the learned motion codes. Removing gloss supervision (\textit{No-gloss}) still provides a moderate improvement over the unaligned variant, reaching sem+kinematic R@1 scores of 75.60 on CSL-Daily and 58.88 on Phoenix-2014T. The full objective further improves performance,($\lambda_{\mathrm{text}} = \lambda_{\mathrm{gloss}} = 0.1$) raising generation to 82.31 on CSL-Daily and 70.25 on Phoenix-2014T.
Increasing to $\lambda = 0.3$ does not consistently improve generation.
It lowers CSL-Daily R@1 to 80.78 while improving Phoenix-2014T R@1 to 74.92.
We use $\lambda=0.1$ by default, which provides a balanced trade-off between semantic alignment and motion reconstruction.

\begin{table}[t]
\centering
\caption{\textbf{Semantic alignment ablation.} 
\textbf{Bold} indicates the best result and \underline{underline} the second best.
}
\label{tab:lambda}
\setlength{\tabcolsep}{3pt}
\resizebox{\columnwidth}{!}{%
\scriptsize
\begin{tabular}{l ccc  ccc}
\toprule
 \multirow{2}{*}{\textbf{Method}} & \multicolumn{3}{c}{\textbf{CSL-Daily}} & \multicolumn{3}{c}{\textbf{Phoenix-2014T}} \\
 \cmidrule(lr){2-4} \cmidrule(lr){5-7}
 & R@1$\uparrow$ & R@5$\uparrow$ & R@10$\uparrow$
 & R@1$\uparrow$ & R@5$\uparrow$ & R@10$\uparrow$ \\
\midrule
$\lambda_\mathrm{text}=0$, $\lambda_\mathrm{gloss}=0$ & 72.96 & 87.24 & 91.75 & 52.96 & 84.27 & 91.12 \\
$\lambda_\mathrm{text}=0.3$, $\lambda_\mathrm{gloss}=0.3$          & \underline{80.78} & \underline{91.16} & \underline{93.96} & \textbf{74.92} & \textbf{92.83} & \textbf{95.79} \\
No-gloss $(\lambda_\mathrm{gloss}=0.0)$    & 75.60 & 89.80 & 93.62  & 58.88 & 84.27 & 90.50 \\
\rowcolor{ours!20}SRVQ (sem+kinematic) & \textbf{82.31} & \textbf{94.13} & \textbf{96.34}  & \underline{70.25} & \underline{87.54} & \underline{92.52} \\
\bottomrule
\end{tabular}
}
\end{table}

\begin{figure}[t]
  \includegraphics[width=0.82\columnwidth]{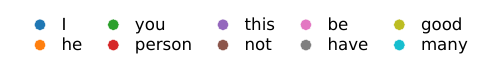}
  \centering
  \begin{subfigure}{0.4\columnwidth}
    \centering
    \includegraphics[width=\linewidth]{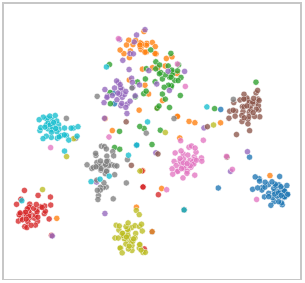}
    \caption{SRVQ level-0}
    
  \end{subfigure}
  \hfill
  \begin{subfigure}{0.4\columnwidth}
    \centering
    \includegraphics[width=\linewidth]{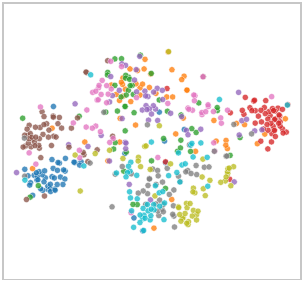}
    \caption{Reconstruction-only~\cite{soke}}
    
  \end{subfigure}
  
  \caption{\textbf{Lexical organization of motion codes.}
  t-SNE visualization of gloss-span motion features on CSL-Daily.
  Colors denote gloss identities. SRVQ level-0 forms clearer lexical neighborhoods than reconstruction-only tokenization.}
  \label{fig:token_manifold}
\end{figure}

\noindent\textbf{Lexical Organization of Semantic Codes.}
We further examine whether the learned semantic codes organize motions according to lexical content, comparing their latent structure with the \textit{reconstruction-only} codes from SOKE~\citep{soke}.
To this end, we select ten frequent glosses from CSL-Daily and extract the token features corresponding to their annotated gloss spans. We visualize the resulting representations with t-SNE~\citep{van2008visualizing}, color-coding samples by their associated gloss. \cref{fig:token_manifold} shows that our text-aligned tokenization form clearer lexical neighborhoods than the reconstruction-only VQ-VAE from SOKE~\citep{soke}. This suggests that reconstruction-only tokenization does not explicitly encourage lexical organization, whereas our semantic alignment objective induces a motion code space that better reflects sign-level content.
The quantitative neighborhood statistics in \cref{tab:token_manifold} confirm the same trend. Our representation increases the nearest-neighbor same-keyword rate (NN) from 56.2\% to 73.8\%, improves both precision-at-$K$ metrics (P@$K$), and raises the silhouette score (Sil.) from 0.032 to 0.260. 
This confirms that semantic supervision shapes level-0 codes toward text-aligned generation.


    
    



\begin{table}[t]
\centering
\caption{\textbf{Quantitative lexical organization of semantic codes.} 
}
\label{tab:token_manifold}
\setlength{\tabcolsep}{3.6pt}
\scriptsize
\begin{tabular}{lcccc}
\toprule
\textbf{Representation} & NN$\uparrow$ & P@5$\uparrow$ & P@10$\uparrow$ & Sil.$\uparrow$ \\
\midrule
Reconstruction-only \citep{soke} & 56.2 & 52.8 & 49.6 & 0.032 \\
\rowcolor{ours!20}SRVQ ($Q_0$) & \textbf{73.8} & \textbf{73.4} & \textbf{73.2} & \textbf{0.260} \\
\bottomrule
\end{tabular}
\end{table}

\noindent\textbf{Text Conditioning Design.} \cref{tab:conditioning} compares conditioning strategies for the full generation pipeline, namely the use of a global sentence embedding \textit{vs.} the word-aware textual representation. Cross-attending both the global textual representation ($\mathbf{e}_{\mathrm{sent}}$) and the word ($\mathbf{W}$) consistently improves over global sentence conditioning alone, \eg, 82.31 \vs 75.34 R@1 in CSL-Daily and 92.52 \vs 79.28 R@10 for Phoenix-2014T, showing the benefit of exposing lexical information to the generators. Notably, the gain is particularly large on Phoenix-2014T, with an approximately 29-point improvement in R@1. We hypothesize that this is due to the Phoenix-2014T’s narrow weather-forecast domain, where sentences share coarse semantics but differ in key lexical cues such as cold, rain, or winter. Thus, global sentence embeddings may fail to capture the exact signs required. In contrast, CSL-Daily’s broader everyday vocabulary offers greater sentence-level variation, where word-level lexical information have diminishing returns.

\begin{table}[t]
\centering
\caption{\textbf{Text conditioning ablation.} 
}
\label{tab:conditioning}
\setlength{\tabcolsep}{3.6pt}
\scriptsize
\begin{tabular}{ cc ccc ccc}
\toprule
 \multirow{2}{*}{$\mathbf{e}_{\mathrm{sent}}$} & \multirow{2}{*}{$\mathbf{W}$} & \multicolumn{3}{c}{\textbf{CSL-Daily}} & \multicolumn{3}{c}{\textbf{Phoenix-2014T}} \\
 \cmidrule(lr){3-5} \cmidrule(lr){6-8}
&& R@1$\uparrow$ & R@5$\uparrow$ & R@10$\uparrow$ & R@1$\uparrow$ & R@5$\uparrow$ & R@10$\uparrow$ \\
\midrule
\checkm{} & \crossm{} & 75.34 & 90.14 & 93.62 & 41.28 & 71.81 & 79.28 \\
\rowcolor{ours!20}\checkm{} & \checkm{} & \textbf{82.31} & \textbf{94.13} & \textbf{96.34} & \textbf{70.25} & \textbf{87.54} & \textbf{92.52} \\
\bottomrule
\end{tabular}
\end{table}

\noindent\textbf{Residual Levels.} \cref{tab:res_depth} investigates how many residual levels are needed for sign-motion quantization. We report retrieval metrics as the main comparison, since geometric metrics are less sensitive to the fine-grained kinematic details that are primarily encoded from level 1 onward. \ourmethod{} remains robust when using fewer residual levels, with the number of residuals having a smaller impact on performance than the SRVQ semantic alignment and word-aware conditioning design. Training fewer residual generators does not consistently improve over the six-level model.
We refer to the Supplementary~\ref{supp:additional-ablations} for additional ablations.

\begin{table}[t]
\centering
\caption{\textbf{Number of residuals ablation.}
$^\dagger$ One-level VQ uses semantic-only decoding.
}
\label{tab:res_depth}
\setlength{\tabcolsep}{3.6pt}
\scriptsize
\begin{tabular}{lccc ccc}
\toprule
 \multirow{2}{*}{\textbf{Depth}} & \multicolumn{3}{c}{\textbf{CSL-Daily}} & \multicolumn{3}{c}{\textbf{Phoenix-2014T}} \\
\cmidrule(lr){2-4} \cmidrule(lr){5-7}
 & R@1$\uparrow$ & R@5$\uparrow$ & R@10$\uparrow$
 & R@1$\uparrow$ & R@5$\uparrow$ & R@10$\uparrow$ \\
\midrule
1$^\dagger$ & 75.77 & 89.54 & 93.28 & 61.21 &83.49 & 88.32 \\
2        & 80.53 & 93.11 & 95.24 & 65.89 & 87.54 & 92.37 \\
4        & \textbf{82.48} & 92.18 & 94.47 & 68.85 & 87.54 & \textbf{92.68} \\
\rowcolor{ours!20}6  & 82.31 & \textbf{94.13} & \textbf{96.34} & \textbf{70.25} & 87.54 & 92.52 \\
\bottomrule
\end{tabular}
\end{table}

\subsection{Analysis of the DTW--Retrieval Discrepancy}
\label{sec:length}


We conduct a diagnostic prefix-truncation experiment on the same generated motions on full test set: for a kept prefix ratio $r$, we evaluate only the first $r$ fraction of generated frames while leaving the reference motion and input text unchanged.
This truncation removes the latter part of the predicted signing sequence, and should therefore reduce generation quality because some sign content no longer exists.
As shown in Tab.~\ref{tab:length_sensitivity}, retrieval and FID follow this expectation.
In contrast, DTW scores improve as the evaluated prefix becomes shorter.
This indicates that DTW reflects geometric alignment under temporal warping, which may not always track sentence-level semantic preservation.

\begin{table}[t]
\centering
\caption{\textbf{Length sensitivity of DTW.} }
\label{tab:length_sensitivity}
\setlength{\tabcolsep}{5.0pt}
\renewcommand{\arraystretch}{1.05}

\resizebox{0.75\linewidth}{!}{%
\begin{tabular}{lcccc}
\toprule
\textbf{Dataset / Ratio} & DTW-B$\downarrow$ & DTW-H$\downarrow$ & FID$\downarrow$ & R@1$\uparrow$ \\
\midrule
\multicolumn{5}{l}{\textbf{CSL-Daily}} \\
\rowcolor{ours!20}\textbf{1.0} & 6.787 & 8.535 & \textbf{0.058} & \textbf{82.31} \\
0.9 & 6.763 & 8.532 & 0.059 & 80.53 \\
0.8 & 6.751 & 8.530 & 0.062 & 74.15 \\
0.7 & 6.613 & 8.408 & 0.080 & 64.37 \\
0.6 & 6.561 & 8.365 & 0.105 & 55.02 \\
0.5 & \textbf{6.534} & \textbf{8.351} & 0.140 & 43.45 \\
\midrule
\multicolumn{5}{l}{\textbf{Phoenix-2014T}} \\
\rowcolor{ours!20}\textbf{1.0} & 5.330 & 6.920 & \textbf{0.059} & \textbf{70.25} \\
0.9 & 5.146 & 6.524 & 0.068 & 65.58 \\
0.8 & 5.031 & 6.394 & 0.089 & 53.12 \\
0.7 & 4.956 & 6.335 & 0.121 & 41.12 \\
0.6 & 4.896 & 6.278 & 0.162 & 28.50 \\
0.5 & \textbf{4.852} & \textbf{6.274} & 0.217 & 16.36 \\
\bottomrule
\end{tabular}%
}
\end{table}





\section{Conclusion}
\label{sec:conclusion}

We presented SRVQ, an SLG framework built around a semantic residual quantization hierarchy in which the coarsest level is jointly trained with a semantic alignment objective.
This design creates a discrete semantic intermediate representation that bridges text and motion, enabling a two-level generation process within a single unified quantizer hierarchy, without separate semantic and reconstruction architectures.
Paired with word-aware conditioning for hierarchical code generation, SRVQ surpasses the state-of-the-art on CSL-Daily and Phoenix-2014T by large margins in retrieval quality while the semantic-only variant improves upon prior baselines on most geometric metrics.

\noindent\textbf{Limitations.}
SRVQ still relies on high-quality SMPL-X motion annotations and frame-level gloss supervision during quantizer training, although no gloss annotations are required at inference.
Future work will study more cross-lingual settings, explicit modeling of non-manual markers, and will incorporate a Deaf-participant human study.


\section*{Acknowledgments}
We acknowledge ISCRA for awarding this project access to the LEONARDO supercomputer, owned by the EuroHPC Joint Undertaking, hosted by CINECA (Italy). 
GI acknowledges financial support from \href{https://www.veasyt.com/}{VEASYT}.
NB acknowledges the support of the EU Horizon Europe projects
ELIAS (No.~101120237) and ELLIOT (No.~101214398), and of the FIS project GUIDANCE (No.~FIS2023-03251). This work was carried out in the Vision and Learning joint laboratory of FBK and UniTN.

{
  \small
  \bibliographystyle{ieeenat_fullname}
  \bibliography{main}
}
\clearpage
\appendix
\section*{Supplementary Material}

This supplement provides additional analyses supporting the main paper. 
Specifically, we first report additional validation results based on back-translation and meaningful pose (\cref{supp:additional-eval}). We then present  ablations  (\cref{supp:additional-ablations}) , including text conditioning, tokenizer reconstruction and residual analysis, gloss-annotation robustness, codebook utilization, and seed sensitivity. Finally, we conclude with a study on the effect of motion length (\cref{supp:motion-length}). 

{\color{black}
\section{Additional Evaluators}
\label{supp:additional-eval}
This section complements the main paper retrieval validation with additional evaluators  relying on back-translation (\cref{supp:additional-back-translation}) and pose retrieval and reconstruction (\cref{supp:additional-meaningful-pose}).
\subsection{Back-translation with SLT Evaluators Adapted to SMPL-X}
\label{supp:additional-back-translation}
To obtain an independent semantic check, we adapt two keypoint-based
sign-language translation systems to accept SMPL-X motion as input. The change
targets the input interface: the original keypoint input is replaced by the
corresponding SMPL-X representation.  Table~\ref{tab:results} reports metrics under the Two-Stream SLT
system~\cite{twostream}, while Table~\ref{tab:signmask_soke_results} relies on MSKA~\cite{mska}.
Both evaluators rank SRVQ above SOKE on every reported BLEU and ROUGE metric,
providing a cross-evaluator semantic check in addition to retrieval.

\begin{table}[htbp]
\centering
\caption{\textbf{Back-translation with the Two-Stream SLT evaluator.}
The Two-Stream Network~\cite{twostream} is adapted from keypoint
input to SMPL-X input.}
\label{tab:results}

\resizebox{\linewidth}{!}{%
\begin{tabular}{lcccc}
\toprule
\textbf{Dataset / Method}
& BLEU-1
& BLEU-2 
& BLEU-4 
& ROUGE \\
\midrule
\multicolumn{5}{l}{\textbf{Phoenix-2014T}} \\
SOKE
& 24.75 
& 13.65 
& 7.05  
& 25.14 \\

\rowcolor{ours!20}
SRVQ (sem+kinematic)
& \textbf{35.92}
& \textbf{23.56}
& \textbf{13.47}
& \textbf{36.05} \\

\midrule
\multicolumn{5}{l}{\textbf{CSL-Daily}} \\
SOKE
& 21.60 
& 11.62 
& 4.33  
& 22.58 \\

\rowcolor{ours!20}
SRVQ (sem+kinematic)
& \textbf{25.81}
& \textbf{14.99}
& \textbf{5.97}
& \textbf{26.93} \\

\bottomrule
\end{tabular}%
}
\end{table}

\begin{table}[htbp]
\centering
\caption{\textbf{Back-translation with the MSKA SLT evaluator.}
MSKA~\cite{mska} is adapted from keypoint input to SMPL-X
input.}
\label{tab:signmask_soke_results}

\resizebox{\linewidth}{!}{%
\begin{tabular}{lcccc}
\toprule
\textbf{Dataset / Method}
& BLEU-1
& BLEU-2
& BLEU-4
& ROUGE\\
\midrule
\multicolumn{5}{l}{\textbf{Phoenix-2014T}} \\
SOKE
& 24.44
& 14.28
& 7.90
& 25.80 \\

\rowcolor{ours!20}
SRVQ (sem+kinematic)
& \textbf{39.36}
& \textbf{27.40}
& \textbf{16.39}
& \textbf{40.96} \\

\midrule
\multicolumn{5}{l}{\textbf{CSL-Daily}} \\
SOKE
& 25.42
& 14.78
& 5.50
& 25.99 \\

\rowcolor{ours!20}
SRVQ (sem+kinematic)
& \textbf{29.49}
& \textbf{18.02}
& \textbf{7.42}
& \textbf{30.38} \\

\bottomrule
\end{tabular}%
}
\end{table}
}

{\color{black}
\subsection{Meaningful Pose Evaluation}
\label{supp:additional-meaningful-pose}
We additionally evaluate our method using the Meaningful Pose toolkit~\citep{meaningful_pose_eval}, which provides complementary pose-retrieval and reconstruction metrics over all joints, body, hands and face. Since the original toolkit does not directly support SMPL-X motion, we adapt it by converting SMPL-X sequences to OpenPose keypoints, resampling them to a fixed length, and performing geometry-only pose retrieval using L2 distance. Unlike the main text--motion retrieval, this is motion-to-motion retrieval: each generated pose sequence is directly compared with a gallery of ground-truth pose sequences using only resampled keypoint geometry. The low absolute R@1 values are therefore expected because retrieving the exact paired motion requires matching signer-, timing-, and articulation-specific geometry; they should not be numerically compared with the text--motion R@1, but only compared across methods under this same protocol. Table~\ref{tab:supp_meaningful_pose} reports results for the sem+kinematic and sem-only variants against SOKE, confirming the trends observed with the text--motion retrieval and geometric metrics in the main paper. Specifically, sem+kinematic achieves the strongest pose-retrieval performance on both datasets for all-joint and hand queries, while sem-only generally yields the best reconstruction. Sem+kinematic also improves over SOKE on Phoenix-2014T and achieves comparable or slightly lower reconstruction performance on CSL-Daily. No learned text--motion, back-translation model is involved in this evaluation.
\begin{table*}[t]
\color{black}
\centering
\caption{\textbf{Meaningful Pose evaluation} on CSL-Daily and Phoenix-2014T. SMPL-X motion sequences are converted to OpenPose keypoints. The official DTW and Aggregated Power Distance metrics reported by the evaluation toolkit correspond to the ``All'' DTW and MJE columns, respectively. Best results on each dataset are shown in \textbf{bold}.}
\label{tab:supp_meaningful_pose}
\footnotesize
\setlength{\tabcolsep}{1.4pt}
\renewcommand{\arraystretch}{0.62}

\resizebox{0.8\textwidth}{!}{%
\begin{tabular}{lccccc|ccccc}
\toprule
\multicolumn{11}{c}{\cellcolor{Gray!7}\textbf{(a) Pose Retrieval}}\\
\midrule
& \multicolumn{5}{c|}{\textbf{All joints}} & \multicolumn{5}{c}{\textbf{Hands}} \\
\cmidrule(lr){2-6}\cmidrule(lr){7-11}
\textbf{Dataset / Method} & R@1$\uparrow$ & R@5$\uparrow$ & R@10$\uparrow$ & MRR$\uparrow$ & MedR$\downarrow$ & R@1$\uparrow$ & R@5$\uparrow$ & R@10$\uparrow$ & MRR$\uparrow$ & MedR$\downarrow$ \\
\midrule
\multicolumn{11}{l}{\textbf{CSL-Daily}} \\
SOKE~\citep{soke} & 1.956 & 5.697 & 9.014 & 0.0472 & 216.0 & 2.126 & 5.527 & 8.844 & 0.0474 & 216.0 \\
\rowcolor{ours!20}
SRVQ (sem+kinematic) & \textbf{2.806} & \textbf{7.823} & \textbf{11.650} & \textbf{0.0607} & \textbf{192.0} & \textbf{2.976} & \textbf{7.908} & \textbf{11.310} & \textbf{0.0603} & \textbf{195.5} \\
\rowcolor{ours!20}
SRVQ (sem-only) & 2.551 & 6.718 & 10.714 & 0.0548 & 202.0 & 2.381 & 6.378 & 9.779 & 0.0533 & 206.5 \\
\midrule
\multicolumn{11}{l}{\textbf{Phoenix-2014T}} \\
SOKE~\citep{soke} & 0.467 & 3.583 & 6.075 & 0.0287 & 174.0 & 0.156 & 3.427 & 5.919 & 0.0250 & 164.5 \\
\rowcolor{ours!20}
SRVQ (sem+kinematic) & \textbf{3.115} & \textbf{9.502} & \textbf{14.019} & \textbf{0.0719} & \textbf{111.0} & \textbf{2.025} & \textbf{8.723} & \textbf{14.486} & \textbf{0.0634} & \textbf{104.5} \\
\rowcolor{ours!20}
SRVQ (sem-only) & 1.558 & 6.386 & 10.125 & 0.0526 & 118.5 & 1.246 & 6.698 & 9.502 & 0.0489 & 125.5 \\
\bottomrule
\end{tabular}
}

\vspace{2pt}

\resizebox{0.8\textwidth}{!}{%
\begin{tabular}{llcccccc}

\toprule
\multicolumn{8}{c}{\cellcolor{Gray!7}\textbf{(b) Pose Reconstruction Error}}\\
\midrule
\textbf{Dataset / Metric} & \textbf{Method} & All & Body & Hands & L-Hand & R-Hand & Face \\
\midrule
\multicolumn{8}{l}{\textbf{CSL-Daily}} \\
DTW$\downarrow$ & SOKE~\citep{soke} & 0.7420 & 0.2428 & 1.8177 & 2.0795 & 1.5560 & 0.2611 \\
\rowcolor{ours!20} & SRVQ (sem+kinematic) & 0.7435 & 0.2422 & 1.8120 & 2.0755 & 1.5485 & 0.2678 \\
\rowcolor{ours!20} & SRVQ (sem-only) & \textbf{0.7014} & \textbf{0.2260} & \textbf{1.7340} & \textbf{1.9755} & \textbf{1.4925} & \textbf{0.2384} \\
\cmidrule(lr){2-8}
MJE$\downarrow$ & SOKE~\citep{soke} & 0.0836 & 0.0274 & 0.2065 & \textbf{0.2469} & 0.1660 & 0.0284 \\
\rowcolor{ours!20} & SRVQ (sem+kinematic) & 0.0864 & 0.0284 & 0.2124 & 0.2552 & 0.1697 & 0.0298 \\
\rowcolor{ours!20} & SRVQ (sem-only) & \textbf{0.0824} & \textbf{0.0268} & \textbf{0.2056} & 0.2479 & \textbf{0.1634} & \textbf{0.0268} \\
\cmidrule(lr){2-8}
MAE$\downarrow$ & SOKE~\citep{soke} & 0.1246 & 0.0410 & 0.3062 & \textbf{0.3643} & 0.2481 & 0.0431 \\
\rowcolor{ours!20} & SRVQ (sem+kinematic) & 0.1287 & 0.0424 & 0.3154 & 0.3766 & 0.2542 & 0.0452 \\
\rowcolor{ours!20} & SRVQ (sem-only) & \textbf{0.1226} & \textbf{0.0400} & \textbf{0.3046} & 0.3644 & \textbf{0.2448} & \textbf{0.0407} \\
\midrule
\multicolumn{8}{l}{\textbf{Phoenix-2014T}} \\
DTW$\downarrow$ & SOKE~\citep{soke} & 0.7089 & 0.2410 & 1.6583 & 1.8333 & 1.4832 & 0.2944 \\
\rowcolor{ours!20} & SRVQ (sem+kinematic) & 0.6446 & 0.2172 & 1.4936 & 1.6659 & 1.3212 & 0.2773 \\
\rowcolor{ours!20} & SRVQ (sem-only) & \textbf{0.5978} & \textbf{0.2004} & \textbf{1.3924} & \textbf{1.4844} & \textbf{1.3003} & \textbf{0.2532} \\
\cmidrule(lr){2-8}
MJE$\downarrow$ & SOKE~\citep{soke} & 0.0838 & 0.0284 & 0.1978 & 0.2219 & 0.1738 & 0.0338 \\
\rowcolor{ours!20} & SRVQ (sem+kinematic) & 0.0778 & 0.0262 & 0.1814 & 0.2020 & 0.1607 & 0.0329 \\
\rowcolor{ours!20} & SRVQ (sem-only) & \textbf{0.0700} & \textbf{0.0236} & \textbf{0.1638} & \textbf{0.1750} & \textbf{0.1526} & \textbf{0.0291} \\
\cmidrule(lr){2-8}
MAE$\downarrow$ & SOKE~\citep{soke} & 0.1217 & 0.0421 & 0.2844 & 0.3143 & 0.2545 & 0.0505 \\
\rowcolor{ours!20} & SRVQ (sem+kinematic) & 0.1143 & 0.0390 & 0.2649 & 0.2914 & 0.2384 & 0.0490 \\
\rowcolor{ours!20} & SRVQ (sem-only) & \textbf{0.1026} & \textbf{0.0349} & \textbf{0.2386} & \textbf{0.2522} & \textbf{0.2249} & \textbf{0.0435} \\
\bottomrule
\end{tabular}
}
\end{table*}
\color{black}

}
\section{Additional Ablations}
\label{supp:additional-ablations}
\subsection{Additional Text-Conditioning Analysis}

\cref{tab:supp_word_only} extends the main text-conditioning comparisons by
evaluating the final word-aware model against a word-token-only variant under the
same generation-validation protocol.  
The word-aware model outperforms the word-token-only variant across all retrieval metrics (+1.44 R@1, +3.06 R@5, +1.19 R@10) on CSL-Daily, consistent with the main-paper conditioning ablation (\cref{tab:conditioning}), where removing word tokens caused a larger drop ($-$6.97 R@1) than removing the global sentence embedding here ($-$1.44), confirming that lexical word-level information is the stronger conditioning signal.

\begin{table}[H]
\centering
\caption{\textbf{Word-token-only comparison on CSL-Daily.}
Both rows use the final generation-validation protocol. DTW-B: DTW-JPE body; DTW-H: DTW-JPE averaged over left and right hands.}
\label{tab:supp_word_only}
\setlength{\tabcolsep}{3.0pt}
\resizebox{\columnwidth}{!}{%
\scriptsize
\begin{tabular}{lcccccc}
\toprule
\textbf{Conditioning} & R@1$\uparrow$ & R@5$\uparrow$ & R@10$\uparrow$ & FID$\downarrow$ & DTW-B$\downarrow$ & DTW-H$\downarrow$ \\
\midrule
\rowcolor{ours!20}Word-aware (ours)
& 82.31 & 94.13 & 96.34 & 0.058 & 6.79 & 8.54 \\
Word tokens only
& 80.87 & 91.07 & 95.15 & 0.057 & 6.83 & 8.63 \\
\bottomrule
\end{tabular}}
\end{table}

{\color{black}
\subsection{Robustness to Gloss Perturbations}
We evaluate robustness to temporal gloss-span noise and gloss dropout under the training protocol. Each row in Table~\ref{tab:supp_gloss_robustness} corresponds to a separately trained variant with the same architecture and optimization settings; only the gloss supervision is perturbed during training. All variants are evaluated on the unmodified test sets, and no gloss annotations are provided at inference.
\begin{table*}[t]
\color{black}
\centering
\caption{Robustness to gloss perturbations. We report sem+kinematic generation for the final model and perturbation settings.}
\label{tab:supp_gloss_robustness}
\setlength{\tabcolsep}{4.0pt}
\renewcommand{\arraystretch}{1.05}
\begin{tabular}{lcccccccc}
\toprule
\multirow{2}{*}{\textbf{Setting}}
& \multicolumn{4}{c}{\textbf{CSL-Daily}}
& \multicolumn{4}{c}{\textbf{Phoenix-2014T}} \\
\cmidrule(lr){2-5}\cmidrule(lr){6-9}
& R@1$\uparrow$ & R@5$\uparrow$ & R@10$\uparrow$ & FID$\downarrow$
& R@1$\uparrow$ & R@5$\uparrow$ & R@10$\uparrow$ & FID$\downarrow$ \\
\midrule
\rowcolor{ours!20}
SRVQ (sem+kinematic)
& \textbf{82.31} & \textbf{94.13} & \textbf{96.34} & 0.0580
& 70.25 & 87.54 & 92.52 & 0.0590 \\
Gloss span shift $\pm$4 frames
& 80.61 & 92.09 & 94.81 & \textbf{0.0488}
& \textbf{70.40} & 90.19 & \textbf{94.86} & \textbf{0.0562} \\
Gloss dropout 20\%
& 77.47 & 91.24 & 94.22 & 0.0495
& 69.78 & \textbf{91.12} & 94.70 & 0.0567 \\
\bottomrule
\end{tabular}
\end{table*}
\color{black}

The final model obtains 82.31 R@1 on CSL-Daily and 70.25 R@1 on Phoenix-2014T. A $\pm4$-frame span shift causes a modest decrease on CSL-Daily (82.31$\rightarrow$80.61), whereas 20\% gloss dropout has a larger effect (82.31$\rightarrow$77.47). Phoenix-2014T R@1 remains nearly unchanged (70.25, 70.40, and 69.78), indicating that the learned representation is relatively tolerant to moderate gloss-boundary noise. The isolated R@5/R@10 increases under perturbation are not consistent across datasets and do not change this R@1 trend.
}

\subsection{Tokenizer Analysis}

\cref{tab:supp_tokenizer} reports reconstruction analysis for the
SRV Quantizer. Full reconstruction encodes the ground-truth motion with all
SRVQ levels and decodes it back to motion, while $Q_0$ reconstruction decodes
only the semantic code level.

\begin{table}[H]
\centering
\caption{\textbf{Tokenizer reconstruction analysis.}
DTW-B: DTW-JPE body; DTW-H: DTW-JPE averaged over left and right hands.}
\label{tab:supp_tokenizer}
\setlength{\tabcolsep}{3.2pt}
\resizebox{\columnwidth}{!}{%
\scriptsize
\begin{tabular}{llcccc}
\toprule
\textbf{Dataset / Method} & R@1$\uparrow$ & FID$\downarrow$ & DTW-B$\downarrow$ & DTW-H$\downarrow$ \\
\midrule
\multicolumn{5}{l}{\textbf{CSL-Daily}}\\
SRVQ full reconstruction & 85.97 & 0.002 & 1.78 & 2.73 \\
SRVQ $Q_0$ reconstruction & 74.49 & 0.018 & 3.33 & 4.81 \\
\midrule
\multicolumn{5}{l}{\textbf{Phoenix-2014T}}\\
SRVQ full reconstruction & 64.49 & 0.008 & 1.56 & 2.89 \\
SRVQ $Q_0$ reconstruction & 40.03 & 0.067 & 3.15 & 5.14 \\
\bottomrule
\end{tabular}}
\end{table}

{\color{black}
\subsection{Residual Code Contribution}
The controlled code-combination analysis in Table~\ref{tab:supp_residual_diagnostic} isolates the contribution of the semantic code and the residual levels without retraining. Each row reports a decoding condition applied to the same test samples and checkpoint. The standard sem+kinematic variant is $\hat{Q}_0+\hat{Q}_{1:L}$; the remaining conditions replace predicted semantic or residual codes with codes extracted from the ground-truth motion and serve as diagnostic oracles.
\begin{table}[t]
\color{black}
\centering
\scriptsize
\setlength{\tabcolsep}{2.5pt}
\renewcommand{\arraystretch}{0.88}
\caption{Controlled residual contribution analysis. $\hat{Q}_{\ell}$ denotes predicted codes at level $\ell$ and $Q^{\mathrm{GT}}_{\ell}$ denotes codes from the ground-truth motion. DTW-B, DTW-H, and PA-H are lower-is-better; hand metrics average the left and right hands.}
\label{tab:supp_residual_diagnostic}
\resizebox{\columnwidth}{!}{%
\begin{tabular}{@{}lrrrrr@{}}
\toprule
\textbf{Dataset / Code combination} & R@1 & R@5 & DTW-B & DTW-H & PA-H \\
\midrule
\multicolumn{6}{l}{\textbf{CSL-Daily}}\\
$\hat{Q}_0+$zero residual & 77.81 & 89.97 & 6.296 & 8.376 & 1.508 \\
$\hat{Q}_0+\hat{Q}_{1:L}$ & 82.31 & 94.13 & 6.787 & 8.535 & 1.544 \\
$\hat{Q}_0+Q^{\mathrm{GT}}_{1:L}$ & 77.98 & 88.95 & 5.140 & 7.519 & 1.513 \\
$Q^{\mathrm{GT}}_0+\hat{Q}_{1:L}$ & 74.15 & 86.90 & 4.945 & 6.430 & 0.845 \\
$Q^{\mathrm{GT}}_0+Q^{\mathrm{GT}}_{1:L}$ & 85.97 & 93.45 & 1.779 & 2.727 & 0.390 \\
\midrule
\multicolumn{6}{l}{\textbf{Phoenix-2014T}}\\
$\hat{Q}_0+$zero residual & 57.48 & 78.97 & 4.700 & 6.410 & 1.262 \\
$\hat{Q}_0+\hat{Q}_{1:L}$ & 70.25 & 87.54 & 5.330 & 6.920 & 1.354 \\
$\hat{Q}_0+Q^{\mathrm{GT}}_{1:L}$ & 60.90 & 82.71 & 3.047 & 4.651 & 1.168 \\
$Q^{\mathrm{GT}}_0+\hat{Q}_{1:L}$ & 53.43 & 78.50 & 4.411 & 6.193 & 0.882 \\
$Q^{\mathrm{GT}}_0+Q^{\mathrm{GT}}_{1:L}$ & 64.49 & 85.05 & 1.564 & 2.889 & 0.406 \\
\bottomrule
\end{tabular}
}
\end{table}
\color{black}

Relative to semantic-only decoding ($\hat{Q}_0+$zero residual), predicted
residuals improve R@1 by 4.50 and 12.77 points on CSL-Daily and Phoenix-2014T,
respectively, while increasing DTW-B by 0.491 and 0.630. Replacing predicted
residuals with ground-truth residuals substantially reduces DTW-B
(6.787$\rightarrow$5.140 and 5.330$\rightarrow$3.047) but lowers retrieval. Thus, residuals improve text-aligned generation, while their predicted realization need not match one signer-specific reference trajectory in timing, coarticulation, or style.  Notably, adding ground-truth residuals $Q^{\mathrm{GT}}_{1:L}$ improves geometric metrics showcasing how reference-specific kinematics allow better reconstruction of the target.
}

\subsection{Lexical Organization on Phoenix-2014T }

\begin{figure}[t]
  \centering
  \includegraphics[width=0.86\columnwidth]{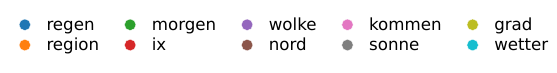}

  \begin{subfigure}{0.48\columnwidth}
    \centering
    \includegraphics[width=\linewidth]{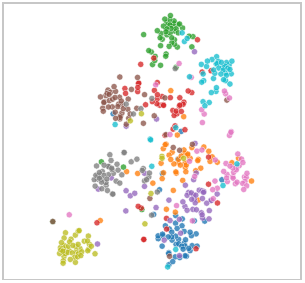}
    \caption{SRVQ level-0}
  \end{subfigure}
  \hfill
  \begin{subfigure}{0.48\columnwidth}
    \centering
    \includegraphics[width=\linewidth]{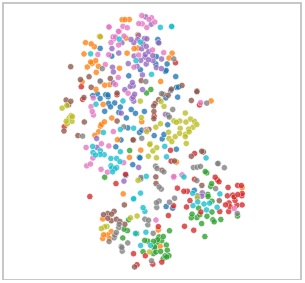}
    \caption{Reconstruction-only~\cite{soke}}
  \end{subfigure}

 \caption{\textbf{Lexical organization of motion codes.}
  t-SNE visualization of gloss-span motion features on Phoenix-2014T.
  Colors denote gloss identities. SRVQ level-0 forms clearer lexical neighborhoods than reconstruction-only tokenization.}
  \label{fig:supp_phx_tsne}
\end{figure}

The main paper visualizes lexical organization on CSL-Daily. We additionally
show the similar analysis on Phoenix-2014T in \cref{fig:supp_phx_tsne}. The
selected glosses are frequent Phoenix-2014T weather-domain signs. 
Despite the limited semantic diversity of the dataset, SRVQ $Q_0$ still forms clear lexical neighborhoods, whereas the reconstruction-only tokenizer exhibits little separation between gloss groups.
The quantitative neighborhood statistics in \cref{tab:supp_phx_manifold} confirm the same trend: SRVQ $Q_0$ improves the nearest-neighbor same-keyword rate from 52.4\% to 69.4\% and the silhouette score from -0.083 to 0.199.

\begin{table*}[t]
\centering
\caption{Per-level codebook utilization of the final SRVQ tokenizer. 
Active codes and normalized perplexity are higher-is-better, while Top-1 usage is lower-is-better.}
\label{tab:codebook_utilization_full}
\setlength{\tabcolsep}{3.6pt}
\renewcommand{\arraystretch}{1.05}

\resizebox{0.8\textwidth}{!}{%
\begin{tabular}{lllcccccc}
\toprule
\textbf{Metric} & Dataset & Part & Q0 & Q1 & Q2 & Q3 & Q4 & Q5 \\
\midrule

\multirow{6}{*}{Active codes (\%) $\uparrow$}
& \multirow{3}{*}{CSL-Daily}
& Body       & 79.17 & 96.88 & 95.83 & 97.92 & 97.92 & 100.00 \\
& & Left hand  & 99.61 & 100.00 & 100.00 & 100.00 & 100.00 & 100.00 \\
& & Right hand & 100.00 & 100.00 & 99.61 & 99.61 & 97.66 & 84.38 \\
\cmidrule(lr){2-9}
& \multirow{3}{*}{Phoenix-2014T}
& Body       & 22.92 & 91.67 & 94.79 & 94.79 & 100.00 & 97.92 \\
& & Left hand  & 98.83 & 99.22 & 98.83 & 99.61 & 100.00 & 99.61 \\
& & Right hand & 99.61 & 99.61 & 100.00 & 100.00 & 98.05 & 91.41 \\

\midrule

\multirow{6}{*}{Norm. perplexity (\%) $\uparrow$}
& \multirow{3}{*}{CSL-Daily}
& Body       & 72.01 & 89.30 & 92.01 & 93.86 & 94.07 & 92.65 \\
& & Left hand  & 68.97 & 68.59 & 68.96 & 67.88 & 69.88 & 67.60 \\
& & Right hand & 79.48 & 86.66 & 92.03 & 95.16 & 92.63 & 72.16 \\
\cmidrule(lr){2-9}
& \multirow{3}{*}{Phoenix-2014T}
& Body       & 22.47 & 87.38 & 87.18 & 86.37 & 84.79 & 84.35 \\
& & Left hand  & 75.63 & 88.37 & 89.48 & 89.49 & 90.57 & 88.39 \\
& & Right hand & 76.00 & 92.61 & 95.67 & 96.67 & 93.21 & 73.83 \\

\midrule

\multirow{6}{*}{Top-1 usage (\%) $\downarrow$}
& \multirow{3}{*}{CSL-Daily}
& Body       & 2.32 & 2.15 & 1.59 & 1.70 & 1.51 & 2.03 \\
& & Left hand  & 2.39 & 2.78 & 2.85 & 3.33 & 3.46 & 4.61 \\
& & Right hand & 1.33 & 1.40 & 0.77 & 0.73 & 0.60 & 0.82 \\
\cmidrule(lr){2-9}
& \multirow{3}{*}{Phoenix-2014T}
& Body       & 6.67 & 2.24 & 2.60 & 4.07 & 3.91 & 3.94 \\
& & Left hand  & 1.73 & 1.20 & 1.32 & 1.50 & 2.05 & 1.91 \\
& & Right hand & 1.56 & 0.84 & 0.76 & 0.68 & 0.76 & 1.17 \\

\bottomrule
\end{tabular}%
}
\end{table*}

\section{Qualitative Sample Metadata}

For completeness, Table~\ref{tab:qualitative_samples} lists the metadata of
the qualitative examples used in the paper and supplementary visualizations,
including the English display text, dataset, sample ID, and original annotated
sentence.

\begin{table*}
\centering
\scriptsize
\setlength{\tabcolsep}{4pt}
\renewcommand{\arraystretch}{1.2}
\caption{\textbf{Qualitative sample metadata.}
The English display text, dataset, sample ID, and original annotated sentence
for the qualitative examples.}
\label{tab:qualitative_samples}
\begin{tabular}{@{}p{0.24\textwidth}p{0.13\textwidth}p{0.23\textwidth}p{0.33\textwidth}@{}}
\toprule
\textbf{English display text}
& \textbf{Dataset}
& \textbf{Sample ID}
& \textbf{Original annotated sentence} \\
\midrule
You're my client, so dinner's on me.
& CSL-Daily
& \texttt{S001423\_P0004\_T00}
& \begin{CJK*}{UTF8}{gbsn}你是我的客户，我应该请你吃饭。\end{CJK*} \\

Exam tomorrow. Bring a pen, no phone.
& CSL-Daily
& \texttt{S001372\_P0008\_T00}
& \begin{CJK*}{UTF8}{gbsn}明天考试，带上笔，不要带手机。\end{CJK*} \\

Today's weather will definitely remain wintry.
& PHOENIX-2014T
& \texttt{26January\_2010\_\allowbreak Tuesday\_heute-105}
& in den nächsten tagen geht es auf jeden fall winterlich weiter immer wieder mal mit etwas schnee . \\
\bottomrule
\end{tabular}
\end{table*}

\begin{table}
\centering
\caption{\textbf{Quantitative evaluation of lexical organization on Phoenix-2014T.}
}
\label{tab:supp_phx_manifold}
\setlength{\tabcolsep}{3.6pt}
\scriptsize
\begin{tabular}{lcccc}
\toprule

\textbf{Representation} & NN$\uparrow$ & P@5$\uparrow$ & P@10$\uparrow$ & Sil.$\uparrow$ \\
\midrule
Reconstruction-only~\citep{soke} & 52.4 & 43.7 & 37.4 & -0.083 \\
\rowcolor{ours!20}SRVQ $Q_0$ & \textbf{69.4} & \textbf{66.9} & \textbf{65.2} & \textbf{0.199} \\
\bottomrule
\end{tabular}
\end{table}

\begin{table}
\centering
\caption{
\textbf{Lexical organization of cumulative and individual SRVQ codes.}
NN, P@5, P@10, and Sil. measure same-gloss neighborhood quality, with higher
values being better. ``Cumulative'' concatenates the codes from $Q_0$ through
the indicated residual level, while ``Individual'' evaluates each residual
level independently.
}
\label{tab:rvq_manifold}

\resizebox{\linewidth}{!}{%
\begin{tabular}{llcccc}
\toprule
\textbf{Dataset / Type}
& Codes
& NN$\uparrow$
& P@5$\uparrow$
& P@10$\uparrow$
& Sil.$\uparrow$ \\
\midrule

\textbf{Phoenix-2014T}\\
Cumulative
& $Q_0$
& 0.6940 & 0.6692 & 0.6520 & 0.1995 \\

& $Q_0$--$Q_1$
& 0.7120 & 0.6888 & 0.6748 & 0.1942 \\

& $Q_0$--$Q_2$
& 0.7160 & 0.7008 & 0.6896 & 0.1984 \\

& $Q_0$--$Q_3$
& 0.7220 & 0.7064 & 0.6880 & 0.1988 \\

& $Q_0$--$Q_4$
& 0.7280 & \textbf{0.7132} & 0.6900 & 0.1999 \\

& $Q_0$--$Q_5$
& \textbf{0.7380} & 0.7116 & \textbf{0.6964} & \textbf{0.2007} \\

\cmidrule(lr){1-6}

Individual
& $Q_1$
& 0.2160 & 0.1848 & 0.1754 & -0.0265 \\

& $Q_2$
& 0.1860 & 0.1692 & 0.1684 & -0.0216 \\

& $Q_3$
& 0.1180 & 0.1124 & 0.1054 & -0.0335 \\

& $Q_4$
& 0.1620 & 0.1428 & 0.1304 & -0.0386 \\

& $Q_5$
& 0.1620 & 0.1608 & 0.1556 & -0.0285 \\

\midrule

\textbf{CSL-Daily}\\
Cumulative
& $Q_0$
& 0.7380 & 0.7344 & 0.7316 & \textbf{0.2596} \\

& $Q_0$--$Q_1$
& 0.7360 & 0.7420 & 0.7354 & 0.2365 \\

& $Q_0$--$Q_2$
& 0.7500 & 0.7432 & 0.7318 & 0.2233 \\

& $Q_0$--$Q_3$
& 0.7580 & 0.7576 & 0.7394 & 0.2148 \\

& $Q_0$--$Q_4$
& \textbf{0.7660} & \textbf{0.7592} & \textbf{0.7430} & 0.2114 \\

& $Q_0$--$Q_5$
& 0.7580 & 0.7560 & 0.7422 & 0.2099 \\

\cmidrule(lr){1-6}

Individual
& $Q_1$
& 0.1300 & 0.1068 & 0.1150 & -0.0351 \\

& $Q_2$
& 0.1320 & 0.1248 & 0.1196 & -0.0341 \\

& $Q_3$
& 0.1260 & 0.1076 & 0.1056 & -0.0352 \\

& $Q_4$
& 0.1020 & 0.1256 & 0.1216 & -0.0350 \\

& $Q_5$
& 0.1220 & 0.1124 & 0.1188 & -0.0349 \\

\bottomrule
\end{tabular}%
}
\end{table}
 Table~\ref{tab:rvq_manifold} separates cumulative
representations from individual residual levels. Cumulative codes preserve
strong same-gloss lexical neighborhoods as residual levels are appended: on
Phoenix-2014T, NN increases from 0.6940 to 0.7380 and P@10 from 0.6520 to
0.6964, while on CSL-Daily P@5 reaches 0.7592 for $Q_0$--$Q_4$. In contrast,
each individual residual level $Q_1$--$Q_5$ has much lower NN/P@$K$ scores and
a negative silhouette score on both datasets. These results indicate that
$Q_0$ is the primary lexical anchor: residual levels preserve the semantic
organization when combined with $Q_0$, but do not independently encode lexical
identity.

\subsection{Codebook Utilization}
On each test set, we report three utilization statistics for each part and RVQ level.
Active is the fraction of codebook entries used by at least one valid token, Norm. perplexity is the code usage perplexity normalized by the codebook size, and Top-1 is the token fraction assigned to the most frequent code.
As shown in Tab.~\ref{tab:codebook_utilization_full}, the hand codebooks are almost fully activated across all levels on both datasets, with high normalized perplexity and low Top-1 usage.
This indicates that SRVQ does not collapse to a few dominant codes.
The only low-activation case is body Q0 on Phoenix-2014T, which is expected for a constrained broadcast domain with limited body pose.
All higher body residual levels remain highly active, showing that the residual hierarchy is still used for kinematic detail.
\subsection{Seed Stability}

\cref{tab:supp_seed} evaluates the final sem+kinematic generator on CSL-Daily
across three inference seeds. Retrieval and geometric metrics show only minor variation, indicating that the reported performance is stable and not dependent on a particular random seed.

\begin{table}[H]
\centering
\caption{\textbf{Seed stability on CSL-Daily.}
All rows use the same sem+kinematic checkpoint. DTW-B: DTW-JPE body; DTW-H: DTW-JPE averaged over left and right hands.}
\label{tab:supp_seed}
\setlength{\tabcolsep}{3.4pt}
\resizebox{\columnwidth}{!}{%
\scriptsize
\begin{tabular}{lcccccc}
\toprule
\textbf{Seed} & R@1$\uparrow$ & R@5$\uparrow$ & R@10$\uparrow$ & FID$\downarrow$ & DTW-B$\downarrow$ & DTW-H$\downarrow$ \\
\midrule
0 & 82.23 & 93.28 & 96.00 & 0.050 & 6.62 & 8.40 \\
1 & 82.91 & 93.96 & 95.92 & 0.050 & 6.60 & 8.41 \\
2 & 83.08 & 94.05 & 96.09 & 0.050 & 6.58 & 8.41 \\
\midrule
Mean $\pm$ std
& $82.74{\pm}0.45$ & $93.76{\pm}0.42$ & $96.00{\pm}0.09$
& $0.050{\pm}0.000$ & $6.60{\pm}0.02$ & $8.41{\pm}0.01$ \\
\bottomrule
\end{tabular}}
\end{table}

\section{Effect of Motion Length}
\label{supp:motion-length}

\cref{tab:supp_length_bucket} reports performance across different ground-truth motion lengths. For retrieval, only the queries are grouped by length, while the gallery remains the full test set, ensuring that retrieval difficulty is unchanged across groups. DTW increases steadily with motion length, whereas retrieval performance does not exhibit a monotonic trend. On Phoenix-2014T, the shortest sequences achieve the lowest DTW scores but the lowest R@1, suggesting that short weather-domain sentences are geometrically easier to align yet semantically more difficult to distinguish.

\begin{table}
\centering
\caption{\textbf{Performance by ground-truth motion length.}
Retrieval is evaluated against the full test gallery. DTW-B: DTW-JPE body; DTW-H: DTW-JPE averaged over left and right hands.}
\label{tab:supp_length_bucket}
\setlength{\tabcolsep}{3.2pt}
\resizebox{\columnwidth}{!}{%
\scriptsize
\begin{tabular}{llcccccc}
\toprule
Dataset & Length & N & R@1$\uparrow$ & R@5$\uparrow$ & R@10$\uparrow$ & DTW-B$\downarrow$ & DTW-H$\downarrow$ \\
\midrule
\multirow{3}{*}{CSL-Daily}
& $\leq$100   & 422 & 82.94 & 94.31 & 96.45 & 4.61 & 5.78 \\
& 101--140    & 367 & 80.11 & 94.28 & 96.46 & 6.62 & 8.32 \\
& $>$140      & 387 & 83.72 & 93.80 & 96.12 & 9.31 & 11.75 \\
\midrule
\multirow{3}{*}{Phoenix-2014T}
& $\leq$76    & 219 & 63.01 & 82.65 & 89.50 & 3.14 & 4.20 \\
& 77--112     & 216 & 75.00 & 89.35 & 95.37 & 5.14 & 6.70 \\
& $>$112      & 207 & 72.95 & 90.82 & 92.75 & 7.85 & 10.03 \\
\bottomrule
\end{tabular}}
\end{table}

\end{document}